\documentclass[onecolumn]{helixworld}

\primarylogo{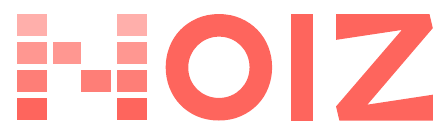}
\productlogo{assets/logos/helixworld.pdf}
\partnerlogos{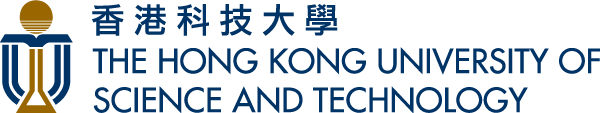=23.5pt}

\usepackage{amsmath,amsfonts,bm}

\def\eqref#1{equation~\ref{#1}}

\def\1{\bm{1}}

\def\vc{{\bm{c}}}

\def\vf{{\bm{f}}}

\def\vh{{\bm{h}}}

\def\vt{{\bm{t}}}

\def\vv{{\bm{v}}}

\def\vx{{\bm{x}}}

\def\mK{{\bm{K}}}

\def\mP{{\bm{P}}}

\def\mR{{\bm{R}}}

\def\mW{{\bm{W}}}

\DeclareMathAlphabet{\mathsfit}{\encodingdefault}{\sfdefault}{m}{sl}
\SetMathAlphabet{\mathsfit}{bold}{\encodingdefault}{\sfdefault}{bx}{n}

\def\gL{{\mathcal{L}}}
\def\gM{{\mathcal{M}}}

\newcommand{\E}{\mathbb{E}}

\newcommand{\R}{\mathbb{R}}

\newcommand{\cmark}{\ding{51}}
\newcommand{\xmark}{\ding{55}}

\hypersetup{%
  hidelinks,
  pdftitle={HelixWorld: A Real-time Interactive Audio-Visual World Model},
  pdfauthor={Lei Ke, Jiahao Pan, Zeyue Tian, Jiaming Wang, Haoyuan Huang, Kam Man Wu, Pengjun Fang, Hongyu Liu, Chenyang Qi, Lin Wang, Ruibin Yuan, Weijia Chen, Fangneng Zhan, Qifeng Chen, Wei Xue, Yike Guo}
}

\title{HelixWorld: A Real-time Interactive Audio-Visual World Model}
\author{%
\textbf{Lei~Ke\textsuperscript{1,2*}, Jiahao~Pan\textsuperscript{1*}, Zeyue~Tian\textsuperscript{1,2*$\dagger$}, Jiaming~Wang\textsuperscript{2}, Haoyuan~Huang\textsuperscript{2},
Kam~Man~Wu\textsuperscript{1,2}}\\[2pt]
\textbf{Pengjun~Fang\textsuperscript{1,2}, Hongyu~Liu\textsuperscript{1}, Chenyang~Qi\textsuperscript{1}, Lin~Wang\textsuperscript{2}, Ruibin~Yuan\textsuperscript{1}}\\[2pt]
\textbf{Weijia~Chen\textsuperscript{2}, Fangneng~Zhan\textsuperscript{1}, Qifeng~Chen\textsuperscript{1}, Wei~Xue\textsuperscript{1$\dagger$}, Yike~Guo\textsuperscript{1}}
}
\affiliations{\textsuperscript{1}The Hong Kong University of Science and Technology\quad \textsuperscript{2}Noiz AI}
\abstract{World simulation is inherently multisensory, demanding synchronized visual and acoustic dynamics in real time. Yet prevailing interactive world models remain strictly silent, focusing exclusively on visual rendering and control while overlooking the acoustic dimension. We present HelixWorld, a real-time interactive audio-visual world model where visual scenes and camera-grounded spatial stereo sound co-evolve natively under user interaction. We curate a high-fidelity spatial audio-visual dataset with true stereo acoustics and metric camera poses, upon which we pre-train a bidirectional teacher conditioned on 6-DoF camera trajectories and user actions. To enable low-latency causal interaction, we distill the teacher into a few-step streaming student via an online trajectory distillation loss, sustaining drift-free joint audio-visual rollouts at 24 FPS on a single GPU. Furthermore, we formalize spatial-acoustic consistency and introduce HelixBench to evaluate whether synthesized sound fields faithfully track dynamic viewpoint motion. Extensive experiments demonstrate that HelixWorld matches state-of-the-art silent world models in visual fidelity and responsiveness, while significantly surpassing existing baselines in camera-aligned spatial-acoustic immersion.
}
\metadata[Project Page]{\textcolor{helixprimary}{\href{https://helixworld.org/}{\underline{\nolinkurl{https://helixworld.org/}}}}}
\metadata[GitHub]{\textcolor{helixprimary}{\href{https://github.com/NoizAI/HelixWorld}{\underline{\nolinkurl{https://github.com/NoizAI/HelixWorld}}}}}

\begin{document}
\maketitle
\begingroup
  \renewcommand{\thefootnote}{\fnsymbol{footnote}}
  \footnotetext[1]{Equal contribution.}
  \footnotetext[2]{Corresponding authors.}
\endgroup
\setcounter{footnote}{0}

\begin{figure}[H]
    \centering
    \includegraphics[width=0.92\linewidth]{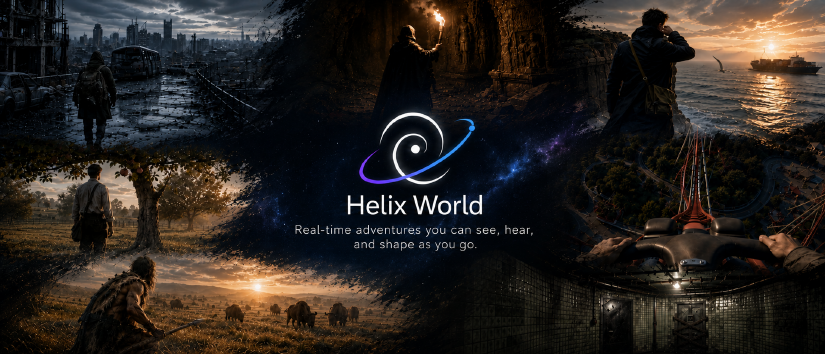}
    \caption{\textbf{HelixWorld.} Real-time interaction across diverse worlds with synchronized video and camera-aligned spatial audio.}
    \label{fig:teaser}
\end{figure}
\clearpage

\justifying
\section{Introduction}
\label{sec:intro}

Interactive world models~\citep{ha2018worldmodels,genie,gamengen,oasis,gamegenx,matrixgame,hunyuangamecraft,worldplay} provide an effective paradigm for physical simulation and virtual environments, allowing users to control generated rollouts in real time via keyboard actions or camera trajectories.
Beyond visual synthesis, physical interactions naturally produce sound that conveys critical cues about surface materials, contact forces, and off-screen events.
To provide immersive simulation, an interactive world model should synthesize visual scenes and spatially grounded stereo audio responsive to user inputs.
However, prevailing world models focus almost exclusively on visual streams, leaving the acoustic dimension largely unexplored.
Despite recent efforts to incorporate audio into world models~\citep{echowm}, achieving fine-grained spatial-acoustic alignment under dynamic camera movement remains an open challenge due to the absence of high-quality spatial audio paired with camera trajectories, severe error accumulation during long-horizon streaming, and instability of causal model training.

Natively coupling interactive visual scenes with spatial audio introduces fundamental challenges that straightforward adaptations of existing models fail to resolve.
Cascaded video-to-audio models~\citep{mmaudio,audiox} lack access to camera trajectories and user actions, leaving sound oblivious to ego-motion, while their sequential latency precludes real-time responsiveness.
Meanwhile, bidirectional audio-visual foundation models~\citep{ltx2,veo3} lack external control interfaces. Moreover, naively converting them into causal autoregressive rollouts leads to catastrophic multimodal drift, where historical error accumulation rapidly degrades both visual stability and acoustic fidelity over extended sequences.

To address these challenges, we present \textbf{HelixWorld}, a real-time interactive audio-visual world model with camera-aligned spatial sound, trained in two stages (see Sec.~\ref{sec:method} for details).
First, \emph{bidirectional teacher training} establishes action controllability and spatial-acoustic fidelity.
We inject continuous 6-DoF camera trajectories into visual attention via PRoPE~\citep{prope} and discrete user actions into diffusion timesteps via AdaLN-Zero~\citep{dit}. Cross-modal attention subsequently binds visual geometry with acoustic features, naturally transferring camera ego-motion into realistic stereo panning.
Second, \emph{causal distillation} converts this bidirectional teacher into a few-step causal student.
To mitigate compounding drift under self-forcing~\citep{selfforcing}, we introduce an \emph{online trajectory distillation loss} that supervises student velocities with teacher Probability Flow ODE targets directly from intermediate rollout states.
Combined with long-horizon streaming fine-tuning, HelixWorld substantially reduces exposure bias and sustains synchronized audio-visual generation at $24$\,FPS on a single NVIDIA H800 GPU.

Beyond model architecture, the successful training of such an interactive system requires large-scale data coupling visual dynamics with high-quality spatial acoustics.
However, existing world model datasets are largely silent; even when audio is present in existing video corpora, it is heavily contaminated by post-production background music, voiceover narration, or pseudo-stereo tracks that provide no authentic spatial cues.
To bridge this training data gap, we construct a scalable, coarse-to-fine curation pipeline structured by computational cost (Sec.~\ref{sec:data}).
The pipeline deploys inter-channel energy probes to eliminate pseudo-stereo audio at ingestion, utilizes multimodal LLMs to purge non-world footage and non-diegetic audio, and automatically recovers frame-accurate metric camera poses paired with decoupled three-track captions (visual scene, foreground acoustic events, and background ambiance), establishing a high-fidelity spatial audio-visual training corpus.

Finally, evaluating an interactive audio-visual world model requires assessing whether synthesized audio physically aligns with dynamic camera motion.
Existing world model benchmarks~\citep{wbench} focus exclusively on visual dynamics while neglecting the acoustic stream, whereas conventional audio-visual benchmarks~\citep{vggsound} evaluate only short, static clips without camera movement.
To close this gap, we introduce \textbf{HelixBench}, the first benchmark dedicated to evaluating interactive audio-visual world models (Sec.~\ref{sec:bench}).
Spanning $1{,}015$ human-verified cases across diverse environments and interaction modes, HelixBench evaluates models across five key dimensions: temporal synchronization, audio-visual semantics, text-audio alignment, acoustic dynamics, and \emph{spatial-acoustic consistency}.
In particular, it formalizes spatial-acoustic consistency by correlating on-screen sound emitter positions with stereo audio panning, directly quantifying whether the synthesized sound field physically rotates with camera ego-motion.

In summary, our main contributions are:

\begin{itemize}
    \item \textbf{HelixWorld Model:} We propose an interactive audio-visual world model that natively unifies continuous 6-DoF camera trajectories and discrete actions, achieving synchronized and camera-responsive rollouts in real time.
    \item \textbf{Causal Distillation Recipe:} We develop a principled distillation framework coupling causal initialization, online trajectory distillation, and streaming fine-tuning, which eliminates compounding rollout drift and sustains real-time $24$\,FPS streaming.
    \item \textbf{Multimodal Data Pipeline:} A systematic multi-stage curation workflow that purges pseudo-stereo and non-diegetic tracks while annotating metric camera extrinsics and decoupled three-track captions, establishing a high-quality spatial corpus.
    \item \textbf{HelixBench:} A benchmark of $1{,}015$ human-verified interactive cases formalizing spatial-acoustic consistency alongside temporal synchrony and multimodal semantics for interactive audio-visual evaluation.
\end{itemize}

\section{Dataset Construction}
\label{sec:data}

Training interactive audio-visual world models requires continuous dynamics, authentic stereo acoustics, and synchronized control.
Yet existing datasets largely lack authentic spatial audio, while in-the-wild video suffers from shot cuts, overlays, and non-diegetic sound.
We address this with an automated pipeline (Fig.~\ref{fig:datapipeline}) that filters multimodal artifacts while annotating metric camera poses, discrete actions, and decoupled captions to power bidirectional pre-training and causal streaming distillation.
More details are provided in Appendix~\ref{app:data}.

\subsection{Data Sources}
\label{sec:data:sources}

\noindent
\begin{minipage}[t]{0.60\linewidth}
\vspace{0pt}
We assemble footage across three complementary sources (Table~\ref{tab:data-mix}).
\emph{Real-world video} captures first-person exploration across diverse indoor and outdoor environments.
\emph{Game footage} records first- and third-person gameplay with native audio, logging ground-truth camera trajectories and control telemetry.
\emph{Open-source corpora} incorporate audio-equipped video subsets~\citep{sekai,gamegenx}, subjected to identical stereo and semantic curation.
\end{minipage}\hfill
\begin{minipage}[t]{0.36\linewidth}
\vspace{-2.5pt}
    \captionsetup{position=top,skip=3pt}
    \centering
    \captionof{table}{\textbf{Corpus by source.}}
    \label{tab:data-mix}
    \small
    \setlength{\tabcolsep}{5pt}
    \renewcommand{\arraystretch}{0.85}%
    \setlength{\aboverulesep}{1pt}%
    \setlength{\belowrulesep}{1pt}%
    \begin{tabular}{@{}lrr@{}}
    \toprule
    Source & Hours & Share \\
    \midrule
    Real-world & 1.8k & 43.9\% \\
    Game & 1.4k & 34.1\% \\
    Open-source & 0.9k & 22.0\% \\
    \midrule
    Total & 4.1k & 100\% \\
    \bottomrule
    \end{tabular}
\end{minipage}\par

\subsection{Curation Pipeline}
\label{sec:data:pipeline}

\begin{figure}[t]
\centering
\includegraphics[width=\linewidth]{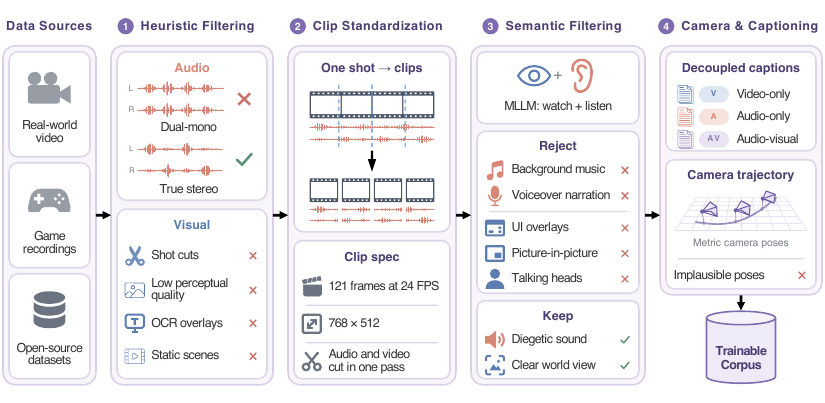}
\caption{\textbf{Data pipeline.} Raw video and audio pass four cost-ordered stages; surviving clips fork into parallel camera recovery and decoupled captioning branches to yield the trainable corpus.}
\label{fig:datapipeline}
\end{figure}

Raw footage cannot be fed directly into training due to noise and artifacts across audio-visual streams.
We design a four-stage pipeline (Fig.~\ref{fig:datapipeline}) ordered by cost, pruning corrupted data before invoking heavy models.

\paragraph{Stage 1: Heuristic filtering.}
Raw videos are split into continuous shots via scene transition detection~\citep{Castellano_PySceneDetect}, discarding fragments under five seconds.
Acoustically, lightweight probes use inter-channel correlation and energy differences to remove mono downmixes and duplicated channels.
Visually, optical character recognition (OCR)~\citep{zhang2026pp} and heuristic detectors prune overlays, banners, and watermarks.
Perceptual quality scoring~\citep{musiq} and motion screening remove severely degraded or frozen clips while preserving static views with physical motion.

\paragraph{Stage 2: Clip standardization.}
Continuous shots are partitioned into fixed-length clips matching the backbone training window.
Video and audio are sliced synchronously without crossing shot boundaries, ensuring temporal alignment.

\paragraph{Stage 3: Semantic filtering.}
To resolve defects requiring high-level scene understanding, an audio-visual multimodal large language model (MLLM)~\citep{qwen3omni} evaluates clips under a structured perception schema.
Visually, it purges non-embodied content (e.g., picture-in-picture, talking heads, occluding overlays) and resolves text ambiguous to OCR.
Acoustically, it strips non-diegetic post-production tracks (background music, voiceover, synthetic effects).
Crucially, sounds are judged by causal origin rather than visibility, preserving off-screen acoustics and natural speech.

\paragraph{Stage 4: Camera and captioning.}
Accepted clips enter two parallel annotation branches.
For web and real-world footage, the camera branch estimates per-frame poses via VGGT-$\Omega$~\citep{vggtomega} and recovers metric scale with Depth Anything V3~\citep{depthanythingv3}, falling back to Metric3D v2~\citep{metric3dv2} when necessary; game footage uses engine-logged ground truth.
After discarding tracking failures and physically implausible trajectories, surviving motions are discretized into an 81-class action vocabulary (Sec.~\ref{sec:action}; benchmarks in Appendix~\ref{app:data}).
Concurrently, the captioning branch synthesizes three decoupled text tracks via independent forward passes: video-only, audio-only, and joint audio-visual descriptions that bind visible sound emitters to their acoustics.
A clip is retained only when both branches succeed.

\noindent
\begin{minipage}[t]{0.55\linewidth}
\vspace{0pt}
In total, this multi-stage pipeline systematically purges visual artifacts and acoustic contamination across the 4.1k raw hours.
Through successive screening and pose verification, 3.0k hours (amounting to 2.1M synchronized clips) survive to form the final trainable corpus, achieving an overall yield of 73.1\% (Table~\ref{tab:datayield}).
The resulting corpus provides a camera-conditioned audio-visual foundation with authentic stereo sound fields and per-frame action annotations, directly powering bidirectional pre-training and causal streaming distillation (Sec.~\ref{sec:method}).
\end{minipage}\hfill
\begin{minipage}[t]{0.41\linewidth}
\vspace{0pt}
    \captionsetup{position=top,skip=5pt}
    \centering
    \captionof{table}{\textbf{Curation yield.}}
    \label{tab:datayield}
    \small
    \begin{tabular}{lrr}
    \toprule
    Stage & Hours & Pass Rate \\
    \midrule
    Raw footage & 4.1k & 100\% \\
    Heuristic filtering & 3.4k & 81.9\% \\
    Semantic filtering & 3.2k & 95.5\% \\
    Camera annotation & 3.0k & 93.5\% \\
    \midrule
    Trainable corpus & 3.0k & 73.1\% \\
    \bottomrule
    \end{tabular}
\end{minipage}\par

\section{Method}
\label{sec:method}

\begin{figure}[t]
\centering

\includegraphics[width=\linewidth]{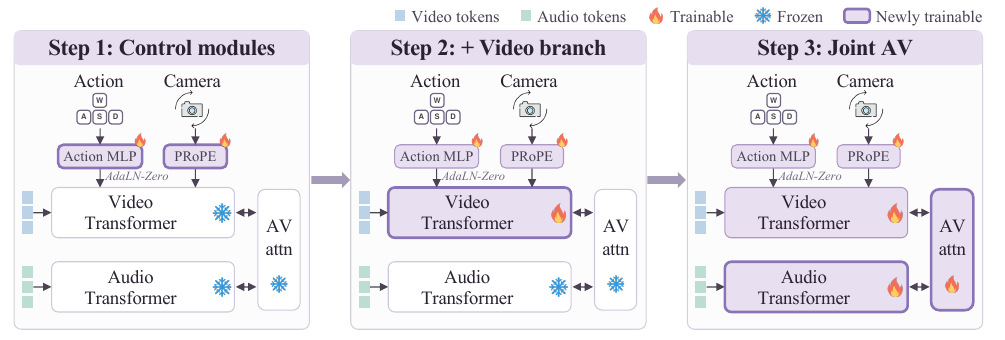}
\caption{\textbf{Progressive condition injection.} Three-stage training: optimize control modules with the backbone frozen, unfreeze the video branch, and jointly train the full audio-visual backbone.}
\label{fig:bidirectional}
\end{figure}

An interactive audio-visual world model requires responsive \emph{action following} (steering visual dynamics and spatial sound) and causal \emph{interactivity} (real-time generation).
To this end, HelixWorld follows a two-stage training pipeline (Figs.~\ref{fig:bidirectional} and~\ref{fig:causal}): Sec.~\ref{sec:bidirectional} trains a bidirectional teacher with progressive camera and action conditioning, while Sec.~\ref{sec:causal} distills it into a few-step causal student.

\subsection{Bidirectional Teacher Training}
\label{sec:bidirectional}

We train the bidirectional model on our corpus (Sec.~\ref{sec:data}) by progressively injecting control signals.
We employ both continuous camera trajectories and discrete user actions as control inputs.
Both controls enter the video branch and propagate to the audio branch through cross-modal attention.

\paragraph{Condition Injection.}
\label{sec:action}
Following~\citet{worldplay}, we condition the video branch on both continuous camera trajectories and discrete actions.
Continuous 6-DoF camera poses are injected via Projective Relative Positional Encoding (PRoPE)~\citep{prope}, transforming visual attention queries and keys by their respective camera projection matrices $\mP_t = \operatorname{diag}(\mK_{f,t}, 1)\mW_t \in \R^{4\times4}$.
Discrete actions are represented by an $81$-class vocabulary over a $9\times9$ grid of translation and view-rotation velocity bins~\citep{hunyuangamecraft,worldplay}.
For each frame $t$, the action index $a_t$ is embedded and added to the diffusion noise level embedding to modulate the transformer backbone via adaptive layer normalization (AdaLN-Zero)~\citep{dit}:
\begin{equation}
\vh_{\mathrm{cond},t} = \mathrm{Emb}_\sigma(\sigma) + \mathrm{MLP}\big(\phi(a_t)\big).
\end{equation}
Both controls enter the video branch and propagate to audio via cross-modal attention.

\paragraph{Progressive training.}
Directly tuning the entire backbone on unaligned controls destabilizes pretrained representations.
We therefore introduce conditioning in three progressive steps.
First, we freeze the backbone and optimize only the control modules (the action MLP and PRoPE projections).
Second, we unfreeze the video branch to adapt visual features to control inputs.
Finally, we unfreeze the full audio-visual backbone for joint flow-matching optimization:
\begin{equation}
\label{eq:joint_fm}
\gL_{\mathrm{FM}} = \gL_{\mathrm{FM}}^{V} + \lambda_A \gL_{\mathrm{FM}}^{A}.
\end{equation}

\subsection{Causal Distillation}
\label{sec:causal}

\begin{figure}[t]
\centering

\includegraphics[width=\linewidth]{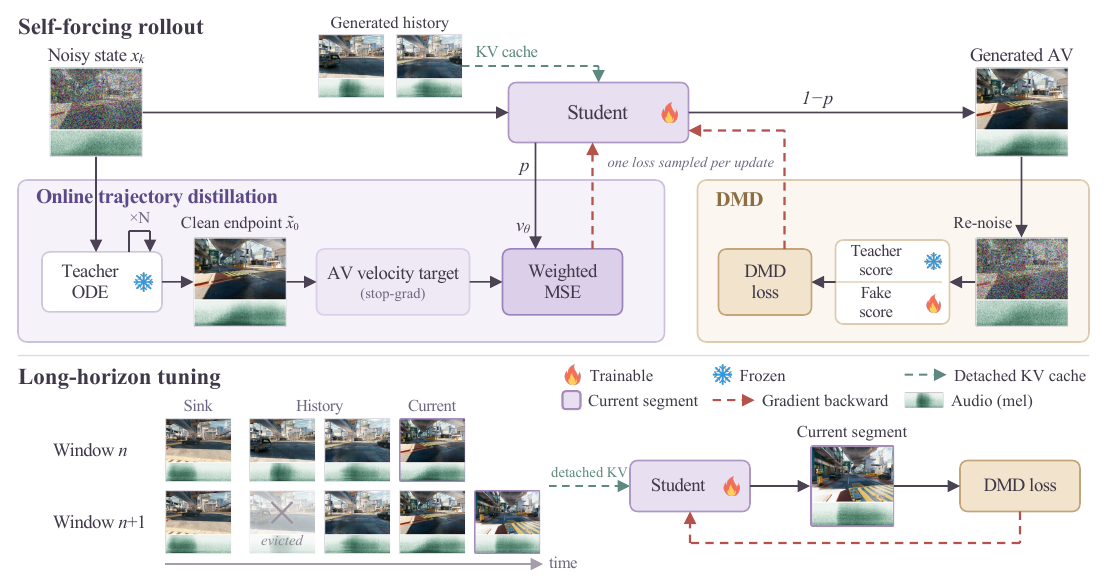}
\caption{\textbf{Causal distillation pipeline.} \emph{Top:} Self-forcing updates stochastically alternate between online trajectory distillation and distribution matching. \emph{Bottom:} Long-horizon tuning extends rollouts segment by segment via a sliding KV cache with initial sink frames and detached history.}
\label{fig:causal}
\end{figure}

To enable real-time generation, we distill the bidirectional teacher into a few-step causal student.
Addressing the visual drift of standard self-forcing (DMD2)~\citep{selfforcing,dmd2}, we propose an \emph{online trajectory distillation loss} that anchors student rollouts to the teacher's probability flow, complemented by long-horizon streaming tuning.

\paragraph{Causal initialization.}
We first synchronize video latents and co-temporal audio tokens into temporal blocks of length $\Delta t$.
Attention within each block is fully bidirectional to maintain audio-visual interactions, while cross-block attention is strictly causal via a sliding Key-Value (KV) cache.
Based on this structure, we adapt the teacher using the joint flow-matching objective $\gL_{\mathrm{FM}}$ (Eq.~\ref{eq:joint_fm}) conditioned on ground-truth history firstly.
To provide a stable starting checkpoint for few-step sampling, we then initialize the student by regressing its single-step predictions onto the Probability Flow ODE endpoints of the causal teacher~\citep{causalforcing}:
\begin{equation}
\label{eq:ode_init}
\gL_{\mathrm{init}}(\theta) = \E_{\vx_\sigma, \sigma, \vc} \left\| \hat{\vx}_0^\theta(\vx_\sigma, \sigma, \vc) - \hat{\vx}_0^{\mathrm{teacher}}(\vx_\sigma, \vc) \right\|_2^2 ,
\end{equation}
where $\vx_\sigma$ is the joint audio-visual latent, and $\vc$ denotes historical context and conditions.

\paragraph{Self-forcing with online trajectory distillation.}
Following causal initialization, the student unrolls blocks autoregressively via self-forcing~\citep{selfforcing} with its KV cache.
Standard self-forcing uses distribution matching ($\gL_{\mathrm{DMD}}$)~\citep{dmd2}, updating the student via score differences between a frozen teacher and a learned fake-score model tracking rollout distributions.
However, distribution matching alone reduces sample diversity~\citep{dp-dmd} and causes long-rollout drift (Fig.~\ref{fig:drift}; Table~\ref{tab:trajectory-diversity}).
To preserve diversity and stabilize rollouts, we introduce an \emph{online trajectory distillation loss} ($\gL_{\mathrm{traj}}$) alongside $\gL_{\mathrm{DMD}}$.

At step $k$ with noise $\sigma_k$ and state $\vx_k = (\vx_k^V, \vx_k^A)$, we integrate the frozen teacher's PF-ODE to $\sigma=0$, yielding endpoint $\tilde{\vx}_0 = (\tilde{\vx}_0^V, \tilde{\vx}_0^A)$.
The student velocity $\vv_\theta = (\vv_\theta^V, \vv_\theta^A)$ is supervised against the implied teacher flow:
\begin{equation}
\label{eq:traj_distill}
\gL_{\mathrm{traj}}(\theta) = \sum_{m\in\{V,A\}} \lambda_m \left\| \vv_\theta^m(\vx_k, \sigma_k) - \operatorname{sg}\!\left( \frac{\vx_k^m - \tilde{\vx}_0^m}{\sigma_k} \right) \right\|_2^2 ,
\end{equation}
where $\operatorname{sg}(\cdot)$ denotes the stop-gradient operator.
Gradients are evaluated strictly at step $k$, with past KV caches and rollout steps detached.
During training, each update stochastically samples $\gL_{\mathrm{traj}}$ with probability $p$ or $\gL_{\mathrm{DMD}}$ with $1-p$.
This routing avoids gradient conflict between objectives, combining distribution matching for sample sharpness with trajectory guidance for rollout stability (see Appendix~\ref{app:training}).

\paragraph{Long-horizon streaming tuning.}
Long rollouts can still suffer from visual degradation and acoustic fading.
To mitigate this drift, we incorporate streaming long tuning~\citep{longlive}.
As shown in Fig.~\ref{fig:causal}, the student unrolls sequences segment-wise with a sliding KV cache, preserving sink frames as anchors while evicting distant history.
Gradients from $\gL_{\mathrm{DMD}}$ are computed only on the active segment with history detached, bounding memory.
\section{HelixBench: A Benchmark for Audio-Visual World Models}
\label{sec:bench}

Existing world-model benchmarks evaluate only silent video generation~\citep{wbench,worldplay}, while audio-visual benchmarks are restricted to static, non-interactive clips~\citep{vggsound}.
In an interactive world model, the generated audio must sound realistic, synchronize with visual content, match text prompts, and adjust its stereo balance as the camera moves.
To evaluate these properties during interactive rollouts, we introduce HelixBench.

\paragraph{Benchmark composition and metrics.}
HelixBench comprises $1{,}015$ human-verified test clips across six scene domains and four subsets: Navigation, Perspective, Event, and Subject.
Each case provides a stereo reference clip, an initial frame, an audio-visual prompt, and continuous camera poses.
Audio quality is measured by KL divergence~\citep{passt} and Fr\'echet Audio Distance (FAD)~\citep{fad} on VGGish embeddings~\citep{vggish}, while multimodal alignment is evaluated via ImageBind~\citep{imagebind} and CLAP~\citep{clap}.
The Event subset measures temporal synchronization via DeSync~\citep{synchformer}, while Perspective evaluates whether stereo audio pans with on-screen sources (Spatial, Appendix~\ref{app:bench}).

\section{Experiments}
\label{sec:exp}
\subsection{Experimental Setup}
\label{sec:exp-setup}

\paragraph{Implementation and training.}
HelixWorld builds on open-source LTX-2.3 and follows the two-stage training procedure in Sec.~\ref{sec:method}. The full implementation details are provided in Appendix~\ref{app:training}.

\paragraph{Baselines.}
To evaluate visual quality, audio fidelity, and audio-visual alignment, we compare HelixWorld against strong baselines~\citep{alayaevoke,echowm,zing,lingbotworld2,worldplay,lyra2,sanawm,dreamxworld,matrixgame3}.
Because EchoWM is the only world model supporting audio-visual generation, we also construct cascaded dubbing baselines using video-to-audio (V2A) models~\mbox{\citep{prismaudio,thinksound,audiox}}.
Specifically, we mute HelixWorld outputs and re-dub them with these models, keeping visual content fixed to isolate audio performance (details in Appendix~\ref{app:bench}).

\subsection{Main Results}
\label{sec:exp-main}

\begin{table}[t]
\caption{\textbf{Visual quality and action following on the WBench navigation split}~\citep{wbench}. Audio indicates native audio synthesis; RTF is measured on a single NVIDIA H800.}
\label{tab:main-visual}
\centering

\resizebox{\linewidth}{!}{%
\begin{tabular}{lcccccccc}
\toprule
Method & Audio & RTF $\downarrow$ & Average $\uparrow$ & Quality $\uparrow$ & Setting $\uparrow$ & Interaction $\uparrow$ & Consistency $\uparrow$ & Physical $\uparrow$ \\
\midrule
Alaya-EVOKE-Turbo & \xmark & 4.29 & 82.0 & 81.9 & 82.1 & 83.9 & 88.1 & 74.0 \\
EchoWM & \cmark & 1.98 & 81.0 & 81.1 & 77.5 & 87.9 & 88.3 & 70.1 \\
Zing-0.5 & \xmark & 2.81 & 81.0 & 80.6 & 77.8 & 84.2 & 88.5 & 73.8 \\
LingBot-World v2 & \xmark & 5.21 & 79.4 & 81.8 & 76.8 & 82.8 & 86.5 & 69.1 \\
HY-World 1.5 & \xmark & 12.33 & 78.1 & 78.1 & 72.2 & 86.8 & 86.9 & 66.3 \\
Lyra 2.0 & \xmark & 30.87 & 76.4 & 77.1 & 73.2 & 85.6 & 79.3 & 66.7 \\
SANA-WM & \xmark & 0.79 & 76.0 & 79.3 & 76.1 & 82.2 & 80.7 & 61.9 \\
DreamX-World & \xmark & 1.04 & 75.0 & 77.5 & 80.8 & 78.6 & 74.9 & 63.3 \\
Matrix-Game 3 & \xmark & 0.58 & 71.3 & 75.5 & 63.6 & 83.6 & 74.5 & 59.3 \\
LTX-2.3 (base) & \cmark & 14.61 & 74.2 & 77.1 & 85.2 & 66.4 & 77.2 & 64.9 \\
\midrule
HelixWorld (ours) & \cmark & 0.77 & 79.9 & 79.2 & 76.5 & 86.4 & 86.4 & 70.9 \\
\bottomrule
\end{tabular}%
}
\end{table}

\begin{table}[t]
\caption{\textbf{Audio-visual quality on HelixBench}. HelixWorld$+$X re-dubs our video using V2A model X; joint generation achieves superior spatial acoustic alignment.}
\label{tab:helixbench-metrics}
\centering
\small
\setlength{\tabcolsep}{4pt}
\begin{tabular}{@{}lcccccc@{}}
\toprule
Model & KL $\downarrow$ & FAD $\downarrow$ & DeSync (s) $\downarrow$ & IB $\uparrow$ & CLAP $\uparrow$ & Spatial $\uparrow$ \\
\midrule
LTX-2.3 (base) & 1.8040 & 7.6354 & \textbf{0.4042} & 0.2777 & 0.2577 & 0.9826 \\
EchoWM & 1.9530 & 7.7472 & 0.6745 & 0.2036 & 0.2352 & 12.6476 \\
HelixWorld + AudioX & 2.0190 & 3.9166 & 1.1764 & 0.2569 & \textbf{0.3094} & $-5.7392$ \\
HelixWorld + ThinkSound & 2.2335 & 6.1876 & 0.4236 & 0.2079 & 0.2888 & 14.8467 \\
HelixWorld + PrismAudio & 2.2427 & 6.3873 & 0.4582 & 0.2152 & 0.2716 & $-9.5072$ \\
\midrule
HelixWorld (bidirectional teacher) & 1.5414 & 3.0775 & 0.4988 & 0.2700 & 0.2676 & 33.2428 \\
HelixWorld (causal student) & \textbf{1.3934} & \textbf{2.3872} & 0.5867 & \textbf{0.2987} & 0.3016 & \textbf{41.7583} \\
\bottomrule
\end{tabular}
\end{table}

\paragraph{Benchmark performance.}
For video quality, we benchmark on the WBench navigation split~\citep{wbench} (Table~\ref{tab:main-visual}).
HelixWorld outperforms most open-source silent baselines across visual quality and control responsiveness, demonstrating faithful action following and physical plausibility.
To assess audio quality and multimodal coherence, we evaluate on HelixBench (Sec.~\ref{sec:bench}) across audio fidelity, temporal sync, semantic alignment, and spatial consistency (Table~\ref{tab:helixbench-metrics}; Appendix~\ref{app:bench}).
HelixWorld surpasses both EchoWM and cascaded dubbing pipelines.
While video-to-audio models achieve reasonable text alignment, they fail to capture emitter locations, resulting in negative spatial scores.
EchoWM retains coarse spatial cues but trails in acoustic quality and semantic binding.
These results demonstrate that HelixWorld effectively grounds spatial acoustics in interactive visual dynamics, achieving physically coherent audio-visual generation.

\paragraph{Inference efficiency.}
On a single NVIDIA H800 GPU, HelixWorld achieves a steady-state RTF of 0.77 at $768{\times}512$ and 24\,fps, including video and audio decoding.
This enables real-time responses to user control (details in Appendix~\ref{app:inference-efficiency}).

\subsection{Ablation Studies}
\label{sec:exp-ablation}

\paragraph{Progressive condition injection.}
To test whether staged conditioning prevents representation collapse (Sec.~\ref{sec:bidirectional}), we compare progressive against full fine-tuning.
Evaluating ViPE-recovered trajectories~\citep{vipe}, progressive fine-tuning reduces both rotation and translation errors (Table~\ref{tab:abl-progressive}).
This confirms that gradually unfreezing branches is essential for learning geometric control while preserving pretrained visual representations.

\paragraph{Multimodal caption composition.}
To test whether explicit sound-source binding improves joint generation, we compare three caption formats (Sec.~\ref{sec:data}): video-only (V), video-plus-audio (V$+$A), and three-track (V$+$A$+$AV).
Cross-evaluating across prompt formats, three-track captions consistently yield the highest ImageBind similarity (Table~\ref{tab:abl-caption}).
This confirms that richer audio-visual descriptions provide direct correspondence cues, strengthening cross-modal consistency.
\begin{table}[!htbp]
\centering
\newcommand{\ablheaderstrut}{\rule[-3.6pt]{0pt}{12pt}}
\newcommand{\ablleftrowstrut}{\rule[-5.4pt]{0pt}{18pt}}
\newcommand{\ablrightrowstrut}{\rule[-3.6pt]{0pt}{12pt}}

\begin{minipage}[t]{0.46\linewidth}
\caption{\textbf{Ablation on camera control.}}
\vspace{1.5pt}
\label{tab:abl-progressive}
\centering
\small
\setlength{\tabcolsep}{3pt}
\renewcommand{\arraystretch}{0}
\begin{tabular*}{\linewidth}{@{\extracolsep{\fill}}lcc@{}}
\toprule
\ablheaderstrut Strategy & Rotation $\downarrow$ & Translation $\downarrow$ \\
\midrule
\ablleftrowstrut Full fine-tuning & 0.1237 & 0.0957 \\
\ablleftrowstrut Progressive fine-tuning & \textbf{0.1144} & \textbf{0.0823} \\
\bottomrule
\end{tabular*}
\end{minipage}\hfill
\begin{minipage}[t]{0.51\linewidth}
\caption{\textbf{Ablation on caption format.}}
\label{tab:abl-caption}
\centering
\small
\setlength{\tabcolsep}{4pt}
\renewcommand{\arraystretch}{0}
\begin{tabular*}{\linewidth}{@{\extracolsep{\fill}}lccc@{}}
\toprule
\ablheaderstrut Train $\backslash$ Infer & V & V$+$A & V$+$A$+$AV \\
\midrule
\ablrightrowstrut V & 0.2372 & 0.2346 & 0.2314 \\
\ablrightrowstrut V$+$A & 0.2366 & 0.2315 & 0.2419 \\
\ablrightrowstrut V$+$A$+$AV & 0.2316 & 0.2323 & \textbf{0.2462} \\
\bottomrule
\end{tabular*}
\end{minipage}
\end{table}

\paragraph{Self-forcing with online trajectory distillation.}
To test whether online trajectory distillation ($\gL_{\mathrm{traj}}$) curbs rollout drift while preserving sample diversity (Sec.~\ref{sec:causal}), we compare self-forcing training with and without $\gL_{\mathrm{traj}}$ on top of DMD.
As shown in Table~\ref{tab:abl-distillation}, incorporating $\gL_{\mathrm{traj}}$ improves overall WBench performance, with marked gains in consistency and physical plausibility.
In extended 30-s rollouts, Figure~\ref{fig:drift} demonstrates that trajectory guidance mitigates visual drift, maintaining cloud structures and scene landmarks with far fewer artifacts.
It also boosts feature diversity on DINOv3 and CLIP over DMD alone (Table~\ref{tab:trajectory-diversity}; Appendix~\ref{app:trajectory}).
These findings validate that anchoring student rollouts to the teacher's probability flow stabilizes autoregressive generation without mode collapse.

\paragraph{Long-horizon streaming tuning.}
To evaluate whether long-horizon tuning sustains generation quality over extended sequences (Sec.~\ref{sec:causal}), we apply it to both self-forcing variants.
As shown in Table~\ref{tab:abl-distillation}, long-horizon tuning delivers consistent, orthogonal gains across both setups, improving overall performance over baselines.
When combined with trajectory distillation, it achieves best overall score.
This confirms that truncated streaming supervision effectively complements block-level trajectory guidance, enabling stable and interactive rollouts over extended horizons.

\begin{table*}[!htbp]
\caption{\textbf{Ablation on self-forcing and long-horizon tuning.} Trajectory distillation ($\gL_{\mathrm{traj}}$) and long-horizon tuning offer complementary gains, jointly reaching the best overall WBench score.}
\label{tab:abl-distillation}
\centering
\small
\setlength{\tabcolsep}{5pt}
\begin{tabular*}{\linewidth}{@{\extracolsep{\fill}}cc cccccc@{}}
\toprule
$\gL_{\mathrm{traj}}$ & Long-Horizon & Average $\uparrow$ & Quality $\uparrow$ & Setting $\uparrow$ & Interaction $\uparrow$ & Consistency $\uparrow$ & Physical $\uparrow$ \\
\midrule
\xmark & \xmark & 77.9 & 77.0 & 75.3 & 85.0 & 85.3 & 66.9 \\
\cmark & \xmark & 78.7 & 77.8 & 74.1 & 86.1 & \textbf{86.6} & \textbf{68.9} \\
\xmark & \cmark & 78.4 & 79.0 & \textbf{77.5} & 86.5 & 83.7 & 65.1 \\
\cmark & \cmark & \textbf{79.1} & \textbf{79.3} & 77.1 & \textbf{86.8} & 84.6 & 67.7 \\
\bottomrule
\end{tabular*}
\vspace{1pt}

\newsavebox{\trajfigurebox}
\newsavebox{\trajtablebox}
\newsavebox{\trajfigcaptionbox}
\newsavebox{\trajtabcaptionbox}
\newlength{\trajrowheight}
\newlength{\trajcaptionheight}
\newlength{\trajtimeheight}
\newlength{\trajpanelwidth}
\setlength{\trajpanelwidth}{0.48\linewidth}

\sbox{\trajfigurebox}{\includegraphics[width=\trajpanelwidth]{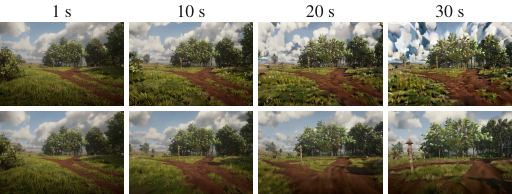}}
\setlength{\trajtimeheight}{0.1184834\ht\trajfigurebox}
\setlength{\trajrowheight}{\dimexpr(\ht\trajfigurebox+\dp\trajfigurebox
  -\trajtimeheight
  -2\heavyrulewidth-\lightrulewidth
  -2\aboverulesep-2\belowrulesep-\abovetopsep-\belowbottomsep)/4\relax}
\newcommand{\trajrowstrut}{\rule[-0.30\trajrowheight]{0pt}{\trajrowheight}}

\sbox{\trajtablebox}{%
\small%
\setlength{\tabcolsep}{2pt}%
\renewcommand{\arraystretch}{0}%
\begin{tabular*}{\trajpanelwidth}{@{\extracolsep{\fill}}lcc@{}}
\toprule
\trajrowstrut Self-forcing & DINOv3 $\uparrow$ & CLIP $\uparrow$ \\
\midrule
\trajrowstrut Without $\gL_{\mathrm{traj}}$ & 0.1089 & 0.0718 \\
\trajrowstrut With $\gL_{\mathrm{traj}}$ & \textbf{0.1326} & \textbf{0.0815} \\
\trajrowstrut Gain & $+21.7\%$ & $+13.5\%$ \\
\bottomrule
\end{tabular*}%
}

\sbox{\trajfigcaptionbox}{%
\begin{minipage}[t]{\trajpanelwidth}
\setlength{\parskip}{0pt}
\setlength{\abovecaptionskip}{0pt}
\setlength{\belowcaptionskip}{0pt}
\captionof{figure}{\raggedright\textbf{Visual quality in 30-s rollouts.}
Self-forcing with $\gL_{\mathrm{traj}}$ (bottom) mitigates visual drift and preserves cloud structures far better than standard DMD (top).}
\label{fig:drift}
\end{minipage}%
}

\sbox{\trajtabcaptionbox}{%
\begin{minipage}[t]{\trajpanelwidth}
\setlength{\parskip}{0pt}
\setlength{\abovecaptionskip}{0pt}
\setlength{\belowcaptionskip}{0pt}
\captionof{table}{\raggedright\textbf{Visual diversity comparison.}
Incorporating online trajectory distillation ($\gL_{\mathrm{traj}}$) into self-forcing improves both DINOv3 and CLIP feature diversity over DMD alone.}
\label{tab:trajectory-diversity}
\end{minipage}%
}

\setlength{\trajcaptionheight}{\dimexpr\ht\trajfigcaptionbox+\dp\trajfigcaptionbox\relax}
\ifdim\dimexpr\ht\trajtabcaptionbox+\dp\trajtabcaptionbox\relax>\trajcaptionheight
  \setlength{\trajcaptionheight}{\dimexpr\ht\trajtabcaptionbox+\dp\trajtabcaptionbox\relax}
\fi

\begin{minipage}[t]{\trajpanelwidth}
\setlength{\parskip}{0pt}\vspace{0pt}\centering
\begin{minipage}[t][\trajcaptionheight][t]{\linewidth}
\vspace{0pt}\usebox{\trajfigcaptionbox}
\end{minipage}\par
\vspace{2.5pt}
\usebox{\trajfigurebox}\par
\end{minipage}\hfill
\begin{minipage}[t]{\trajpanelwidth}
\setlength{\parskip}{0pt}\vspace{0pt}\centering
\begin{minipage}[t][\trajcaptionheight][t]{\linewidth}
\vspace{0pt}\usebox{\trajtabcaptionbox}
\end{minipage}\par
\vspace{2.5pt}
\vspace{\trajtimeheight}
\raisebox{\dp\trajtablebox}{\usebox{\trajtablebox}}\par
\end{minipage}
\end{table*}

\subsection{Qualitative Analysis and User Study}
\label{sec:exp-qual}

\begin{figure}[t]
\centering
\includegraphics[width=\linewidth]{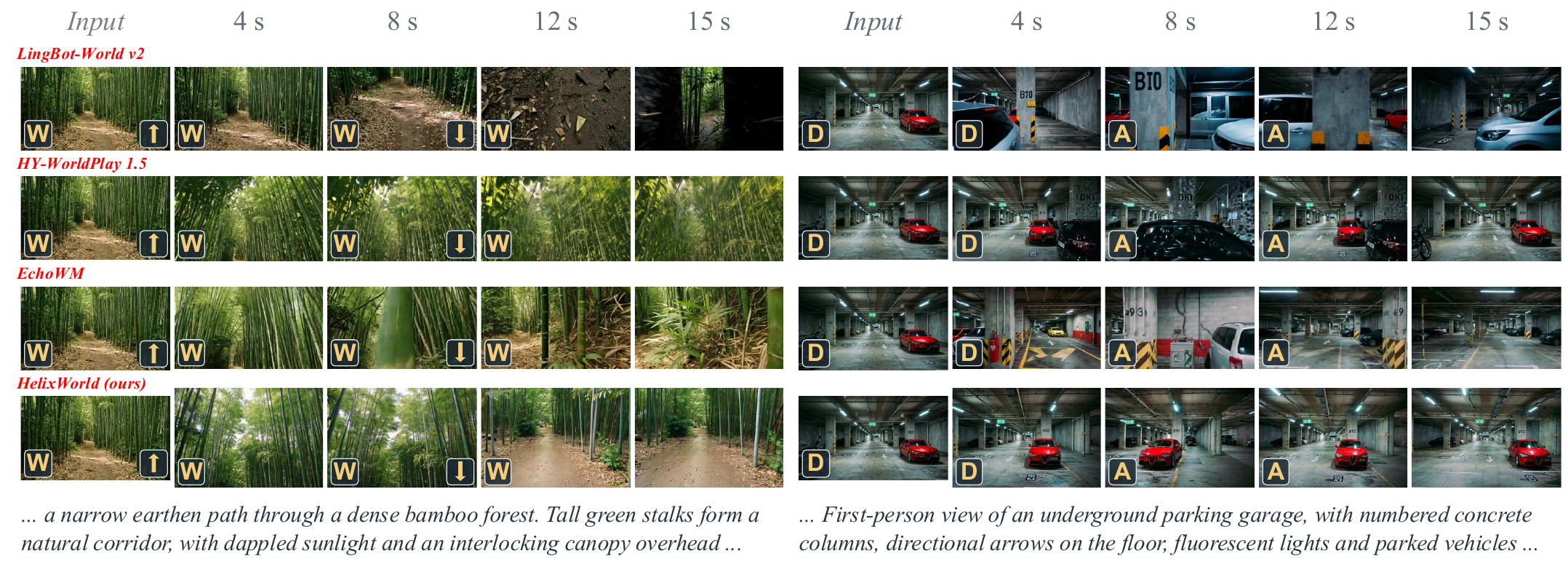}
\caption{\textbf{Visual comparison under interactive control.} HelixWorld retains scene structure during navigation (left) and motion reversal (right). Key overlays indicate controls.}
\label{fig:qualitative}
\end{figure}

\begin{figure}[t]
  \centering
  \setlength{\abovecaptionskip}{6pt}
  \begin{minipage}[t]{0.40\linewidth}
    \setlength{\parskip}{0pt}
    \centering
    \includegraphics[width=\linewidth]{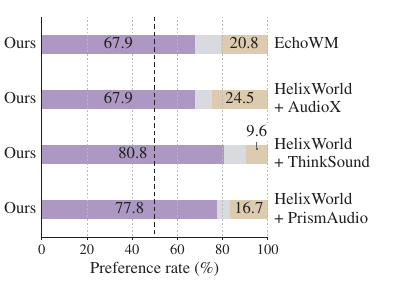}\par
    \vspace{2pt}
    {\small (a) With audio\par}
  \end{minipage}\hspace{0.08\linewidth}
  \begin{minipage}[t]{0.40\linewidth}
    \setlength{\parskip}{0pt}
    \centering
    \includegraphics[width=\linewidth]{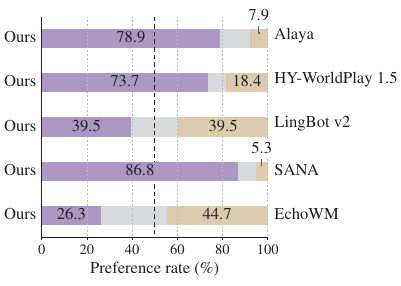}\par
    \vspace{2pt}
    {\small (b) Muted video\par}
  \end{minipage}
  \vspace{1pt}
  \caption{\textbf{User-study preferences.} Purple indicates preference for HelixWorld, gray a tie, and beige preference for the baseline across full audio-visual rollouts (a) and muted video dynamics (b). Dashed vertical lines mark 50\%.}
  \label{fig:user-study}
\end{figure}

\paragraph{Qualitative results.}
Figure~\ref{fig:qualitative} compares HelixWorld with LingBot-World~v2, HY-WorldPlay~1.5, and EchoWM on same scenes with shared first-frame and control inputs.
In the bamboo scene, our model retains the forest corridor as the view rises into the canopy and returns to the path.
In the garage, the red car and nearby pillars remain recognizable through the right-to-left movement reversal.
Additional audio--visual comparisons are provided in Appendix~\ref{app:av-qualitative}.

\paragraph{User study.}

We conduct a blind pairwise user study across audio-visual and muted settings (Figure~\ref{fig:user-study}; Appendix~\ref{app:user}).
With audio, HelixWorld is preferred over all baselines by a clear margin, winning the majority of votes in every matchup.
In muted evaluations, it outperforms Alaya, HY-WorldPlay, and SANA, ties with LingBot~v2, and trails only EchoWM.

\section{Related Work}
\label{sec:related}

\paragraph{Interactive world models.}
Interactive world models~\citep{ha2018worldmodels} simulate future observations under diverse controls, including latent codes~\citep{genie,genie2}, keyboard actions~\citep{gamengen,oasis,gamegenx,matrixgame,hunyuangamecraft}, 6-DoF camera trajectories~\citep{cameractrl,prope}, and hybrid inputs~\citep{worldplay}.
Yet, prevailing models remain silent; while EchoWM~\citep{echowm} incorporates audio, it lacks camera-conditioned spatial acoustics.
HelixWorld conditions a joint audio-visual backbone on continuous camera trajectories and discrete actions, steering viewpoints alongside responsive spatial acoustics.

\paragraph{Audio-visual generation.}
Cascaded systems dub pre-rendered video via video-to-audio synthesis~\citep{mmaudio,moviegen}, but post-hoc dubbing lacks camera extrinsics or control intents, and compounding latency prevents real-time interaction.
In contrast, joint models~\citep{ltx2,veo3} co-denoise video and audio tokens in a shared transformer for cross-modal synchrony.
Yet, they remain passive and bidirectional, operating on fixed-length clips without control interfaces or streaming inference.
We bring interactive control and causal streaming to joint models.

\paragraph{Causal distillation.}
Deploying diffusion models interactively requires converting bidirectional generation into causal streaming.
Recent video distillation schemes mitigate exposure bias through distribution matching (DMD)~\citep{causvid,dmd2}, on-policy KV rollouts (Self-Forcing)~\citep{selfforcing}, ODE regression~\citep{causalforcing}, and long-horizon streaming tuning~\citep{longlive}.
However, these techniques target video alone.
When extended to joint audio-visual diffusion, distribution matching suffers from mode collapse, causing color degradation and acoustic dropouts over extended rollouts.
We stabilize self-forcing with an online trajectory distillation loss, anchoring the student's joint velocity field against teacher probability flows.

\paragraph{Audio-visual evaluation.}
World-model benchmarks such as WBench~\citep{wbench} assess action-conditioned video dynamics but ignore audio entirely.
Conversely, audio-visual benchmarks~\citep{vggsound} measure semantic correspondence (ImageBind~\citep{imagebind}, CLAP~\citep{clap}) and onset synchrony (Synchformer~\citep{synchformer}, AV-Align~\citep{tempotokens}) on short, fixed-viewpoint clips.
While these metrics evaluate \emph{what} sounds occur and \emph{when}, they cannot test \emph{where} sounds originate relative to a moving listener under viewpoint motion.
HelixBench addresses this gap, formalizing \emph{spatial-acoustic consistency} alongside temporal, semantic, and dynamic metrics for interactive audio-visual evaluation.

\section{Conclusion}
\label{sec:conclusion}

In this work, we present HelixWorld, a real-time interactive audio-visual world model where visual scenes and spatial stereo soundscapes co-evolve natively under dynamic user control.
Coupled with a compute-prioritized data curation pipeline, our causal distillation framework suppresses compounding rollout drift to sustain synchronized $24$\,FPS streaming on a single GPU.
Finally, we formalize spatial-acoustic consistency as a foundational criterion separating interactive world models from passive generation, and introduce HelixBench to systematically evaluate it.

\clearpage
\bibliographystyle{plainnat}
\bibliography{helixworld}

\appendix
\clearpage
\section{Architecture and Conditioning Details}
\label{app:arch}

\paragraph{Camera conditioning.}
The camera branch supplements the video self-attention in Sec.~\ref{sec:action}.
Intrinsics are normalized to the latent grid, and extrinsics are expressed relative to the first frame, whose relative pose is the identity.
The focal-only lift retains focal lengths but omits the principal point.
Invalid or missing camera tokens, as well as audio tokens, receive identity transforms.
The camera branch shares the base attention's Q/K/V projections and contributes through a zero-initialized output projection.

\paragraph{Action conditioning.}
The $81$ action classes factor as $\mathrm{id}=9\,\mathrm{translation}+\mathrm{rotation}$.
Translation and rotation each enumerate nine bins in $\{-,0,+\}^2$, for local $(x,z)$ translation and (yaw, pitch), respectively.
Each video latent frame has its own action index.
The index is encoded with $256$-dimensional sinusoidal features and a Linear--SiLU--Linear MLP with a zero-initialized output, then added to the noise-level embedding.
Camera and action controls enter the video branch and reach audio through cross-modal attention.

\section{Training and Streaming Inference Details}
\label{app:training}

\subsection{Temporal Blocks and First-Frame Conditioning}
\label{app:blocks}

A short self-forcing window contains $16$ video latent frames and $127$ audio tokens.
The video is partitioned into four blocks of four latent frames; the corresponding audio blocks contain $27$, $33$, $33$, and $34$ tokens, assigned by their timestamps.
The conditioning image occupies the first latent of the first block.
It participates in joint audio--visual attention at noise level zero, is restored after each denoising and re-noising operation, and is excluded from the generated-token loss.
The remaining $15$ video latents and all audio tokens are generated.

Attention is bidirectional within a block and causal across blocks.
We detach historical latents and committed KV tensors and retain gradients within the current block.
Camera and action sequences are aligned on the source timeline before temporal blocks are selected.

\subsection{Causal Initialization}

\paragraph{Teacher forcing.}
We adapt the bidirectional model to block-causal attention using clean ground-truth history and the joint audio--visual flow-matching objective.
Short clips supply four consecutive blocks.
For long sources, the initialization recipe packs the first block with a later contiguous three-block window and supervises the final target block.
This exposes the model to later source positions within a bounded context.

\paragraph{ODE initialization.}
The student regresses onto precomputed endpoints of the teacher-forced causal model (Eq.~\ref{eq:ode_init}).
The teacher uses $50$ dense solver steps, CFG $3$, and no STG.
Input states are sampled from the four nonzero student knots $\{1.0,0.9,0.7,0.4\}$.
The image-conditioned initialization uses the first frame on every example and optimizes the joint video/audio endpoint MSE with a constant learning rate of $2\times10^{-6}$.

\subsection{Self-Forcing Optimization}
\label{app:sf-optimization}

The generator starts from the ODE-initialized causal model.
The frozen real-score teacher and the trainable fake-score model both start from the bidirectional model after progressive condition injection.
Generator and fake-score adaptation use LoRA with the shared settings in Table~\ref{tab:sf-hyperparameters}.

\paragraph{Rollout and gradient path.}
The student denoises each block with the schedule $[1.0,0.9,0.7,0.4,0]$, drawing independent video and audio noise for each re-noising transition.
A clean, detached KV commit follows each completed block, and later blocks append to this history.
A short training window contains four blocks, all retained in the history context.
The loss covers generated tokens across the window, with gradients retained at one selected denoising step per block.

\paragraph{DMD score evaluation.}
The bidirectional real and fake models evaluate the same noised student sample at the same physical noise level.
For the self-forcing recipe, a sampled raw level $u$ is shifted once as
\begin{equation}
\sigma=\frac{\gamma u}{1+(\gamma-1)u},\qquad \gamma=3.
\label{eq:app-score-shift}
\end{equation}
Generator-score queries use $u\sim\mathcal U(0.02,0.98)$; fake-score training uses $u\sim\mathcal U(0,1)$.
The resulting $\sigma$ is shared by video and audio and used consistently for noising, model queries, and clean-sample conversion.
Student rollouts use the fixed four-knot schedule above.

\paragraph{Online trajectory distillation.}
With probability $0.1$, a generator event uses Eq.~\ref{eq:traj_distill} in place of a DMD update; otherwise it follows five fake-score updates and one DMD generator update.
For a trajectory event, one student denoising index is shared across the four blocks.
The frozen bidirectional teacher integrates from the corresponding noisy state to a detached endpoint on a dense schedule with eight steps per student interval ($32$ over the full schedule).
Gradients update the selected student predictions while conditioning-image tokens and historical KV states remain fixed.

\begin{table}[tbp]
\caption{\textbf{Self-forcing hyperparameters.} Settings shared by DMD-only self-forcing and self-forcing with online trajectory distillation loss.}
\label{tab:sf-hyperparameters}
\centering\small
\setlength{\tabcolsep}{5pt}
\renewcommand{\arraystretch}{1.08}
\begin{tabularx}{\linewidth}{@{}lX@{}}
\toprule
Setting & Value \\
\midrule
Generator / fake-score adaptation & LoRA rank $256$, alpha $256$, dropout $0$ \\
Optimizer & AdamW, $\beta_1=0.9$, $\beta_2=0.999$ \\
Generator / fake learning rate & $10^{-5}$, constant; no warmup or decay \\
Gradient norm clipping & $1.0$ \\
Precision & BF16 with gradient checkpointing \\
Global batch size & $48$; one sample per GPU, no accumulation \\
Training hardware & Six nodes, each with eight H200 GPUs \\
Fake updates per DMD generator event & $5$ \\
Teacher video / audio CFG & $4$ / $2$ \\
Teacher STG & Scale $1$, transformer block $29$ \\
Teacher guidance rescale / modality scale & $0$ / $1$ \\
Student inference CFG / STG & $1$ / $0$ \\
\bottomrule
\end{tabularx}
\end{table}

\subsection{Long-Horizon Tuning and Streaming Inference}
\label{app:streaming}

\paragraph{Long-horizon tuning.}
We apply long-horizon tuning to self-forcing models trained with and without online trajectory distillation, starting from their respective generators and fake-score models.
The student rolls out approximately one minute of synchronized video and audio, advancing through the source in temporal order.
In both variants, the DMD loss supervises the newest window, with earlier latents and KV states detached.
The bidirectional score models process short windows of $16$ video latents and $127$ audio tokens.
Table~\ref{tab:abl-distillation} compares both variants before and after long-horizon tuning.

\paragraph{Bounded inference history.}
The inference context contains the fixed first block, the two most recent completed blocks, and the current target block: at most four blocks in total.
The first block retains the conditioning-image latent.
For example, the fifth target block reads blocks $1$, $3$, and $4$.
Temporal keys are cached before RoPE; video and audio positions share a compact local timeline for the retained context.

\paragraph{Cache reconstruction.}
The reported rebuild evaluations recompute history KV from the selected clean, generated latents before denoising the next target.
Each history block attends to itself and earlier retained blocks under the block-causal mask.
Reconstruction updates cached keys and values while keeping the generated history latents fixed.
The visual quality and diversity comparisons use the same rebuild policy in both arms.
Inference jointly generates both modalities with four stochastic denoising steps, CFG $1$, and STG $0$.

\section{Dataset Construction and Curation Details}
\label{app:data}

This section provides comprehensive details on the data curation and annotation pipeline described in Sec.~\ref{sec:data}, including domain distribution and ethical privacy considerations, signal-level thresholds, MLLM perception schema, camera pose and metric scale recovery, and multimodal caption contracts.

\subsection{Data Sources, Domain Distribution, and Ethical Considerations}
\label{app:data:sources}

We assemble raw footage totaling $4.1$\,k hours across three complementary domains (Table~\ref{tab:data-sources}), systematically balancing real-world dynamics, interactive gaming environments, and diverse open-source distributions.

\paragraph{Real-world web video.}
We collect $1.8$\,k hours of high-resolution first-person exploration, urban driving, and environmental walkthroughs from publicly available video platforms.
To safeguard personal privacy and prevent unauthorized disclosure, we apply automated face de-identification and license-plate obfuscation during pre-processing; any footage centered on identifiable individuals or private personal spaces is systematically purged in Stage~1 and Stage~3.

\paragraph{Game screen recordings.}
We record $1.4$\,k hours of high-dynamic first- and third-person navigation across diverse 3D virtual environments and modern game engines, capturing native multi-channel engine audio and synchronized camera trajectories.
Direct access to engine telemetry provides noise-free ground-truth camera extrinsics and user control inputs, serving as a reliable geometric anchor for interactive control learning.

\paragraph{Open-source corpora.}
We incorporate $0.9$\,k hours from open-domain video corpora with native audio tracks, including Sekai~\citep{sekai} and GameGen-X~\citep{gamegenx}, covering diverse open-world scenes and interactive scenarios.
These subsets are subjected to identical stereo and semantic curation pipelines to ensure corpus-wide distribution uniformity.

\begin{table}[!htbp]
\caption{\textbf{Data sources, domain distribution, and corpus statistics.} Pose origin indicates whether camera trajectories are logged directly from engine states or reconstructed offline via geometric estimation. All statistics reflect the frozen corpus.}
\label{tab:data-sources}
\centering
\small
\setlength{\tabcolsep}{8pt}
\renewcommand{\arraystretch}{1.2}
\begin{tabular}{@{}llrrr@{}}
\toprule
Source Family & Pose Origin & Raw Hours & Share & Trainable Clips \\
\midrule
Real-world web video & Estimated (VGGT-$\Omega$ + DA3) & $1.8$\,k & $43.9\%$ & $\sim$$0.91$\,M \\
Game screen recordings & Engine-logged (ground truth) & $1.4$\,k & $34.1\%$ & $\sim$$0.73$\,M \\
Open-source subsets & Mixed (logged / estimated) & $0.9$\,k & $22.0\%$ & $\sim$$0.46$\,M \\
\midrule
Total Raw Footage & --- & $4.1$\,k & $100\%$ & $\sim$$2.93$\,M \\
\textbf{Trainable Corpus} & Verified metric poses & $\mathbf{3.0}$\,k & $\mathbf{73.1\%}$ & $\mathbf{2.10}$\,M \\
\bottomrule
\end{tabular}
\end{table}

\subsection{Signal-Level Filtering and Clip Standardization (Stages 1 \& 2)}
\label{app:data:signal}

Raw footage exhibits pervasive distribution corruption, including dual-mono downmixes, rapid montage cuts, static title cards, and interface clutter.
Stage~1 executes lightweight signal probes to eliminate defective footage prior to compute-intensive neural processing.

\paragraph{Container and format validation.}
The ingestion probe enforces strict container constraints: video streams must have a spatial resolution of at least $768{\times}512$ with horizontal landscape aspect ratio, and frame rates bounded within $[23.9, 121]$\,fps (subsequently normalized to $24$\,fps).
The audio stream must contain a valid two-channel stereo track, immediately pruning mono downmixes and missing audio streams.
Clips failing container integrity or exhibiting corrupt stream headers are pruned immediately.

\paragraph{True stereo verification.}
A significant fraction of online videos duplicate a single mono recording across two channels, creating ``dual-mono'' tracks devoid of spatial acoustic information.
We measure inter-channel stereo energy over sliding $20$-second analysis windows via normalized root-mean-square difference:
\begin{equation}
\label{eq:stereo_metric}
\delta = \frac{\|L - R\|_2^2}{\|L\|_2^2 + \|R\|_2^2} ,
\end{equation}
where $L$ and $R$ denote the discrete time-domain signals of the left and right channels, respectively.
For identical channels, $\delta \equiv 0$, whereas authentic spatial acoustics yield $\delta \in [0.05, 1.0]$.
We discard any upload with $\delta < 0.01$.
Surviving streams are subsequently verified for phase correlation to eliminate artificial out-of-phase stereo widening artifacts.

\paragraph{Shot transition and temporal continuity.}
We employ PySceneDetect~\citep{Castellano_PySceneDetect} to segment continuous shots, tracking HSV color histogram differences across frames subsampled at $4$\,fps.
A cut boundary is placed whenever the histogram difference exceeds $0.35$.
Fragmented shots shorter than $5$\,seconds are dropped to prevent abrupt scene cuts from contaminating the temporal attention window.
Furthermore, any video whose mean continuous shot duration falls below $2$\,seconds (e.g., promotional trailers or rapid montages) is rejected in its entirety.

\paragraph{Static scene discrimination.}
To eliminate static still-image slideshows and frozen streams while retaining stationary cameras that observe active physical events (e.g., vehicles driving past a fixed camera), we compute the Median Absolute Difference (MAD) between consecutive grayscale frames sampled at $1$\,fps:
\begin{equation}
\mathrm{MAD} = \operatorname{median}_{(x,y)} \big| I_{t+1}(x,y) - I_t(x,y) \big| .
\end{equation}
Shots with $\mathrm{MAD} < 1.5$ (on an 8-bit scale $[0, 255]$) are rejected.
Employing the median rather than the mean prevents localized high-frequency perturbations (such as blinking watermarks, subtitle updates, or compression noise) from falsely validating an otherwise frozen scene.

\paragraph{Clip standardization (Stage 2).}
Surviving continuous shots are segmented into standardized clips matching the backbone training context: exactly $121$ video frames at $24$\,fps ($5.04$\,s) at $768{\times}512$ resolution, coupled with synchronized two-channel $48$\,kHz stereo audio ($241{,}920$ audio samples per clip).
Shots are tiled front-to-back without overlap.
To maximize data efficiency, whenever a trailing segment exceeds $2.5$\,s, an additional clip is carved backwards from the shot terminus, raising temporal coverage from $98.4\%$ to $100\%$.
Audio and video streams are sliced synchronously in a single ffmpeg pass, ensuring microsecond-level temporal synchronization by construction.

\subsection{Semantic Filtering with Audio-Visual MLLM (Stage 3)}
\label{app:data:semantic}

While low-level heuristic probes eliminate signal defects, they cannot reason about high-level multimodal semantics: pixel differences fail to separate 3D physical environments from 2D screen recordings or talking heads, and audio format checks cannot detect added background music.
Stage~3 deploys an audio-visual foundation model (Qwen3-Omni-30B~\citep{qwen3omni}) to evaluate video and audio synchronously under constrained grammar decoding.
A clip is retained if and only if it satisfies two physical criteria:
\begin{itemize}[leftmargin=1.5em,itemsep=1pt,topsep=1pt]
\item \textbf{Visual scene integrity}: The clip must depict an authentic 3D physical environment undergoing dynamic physical evolution, strictly excluding non-world content (e.g., talking-head interviews, desktop screencasts, picture-in-picture feeds) and intrusive graphic or subtitle overlays.
\item \textbf{Acoustic authenticity}: The clip must feature natural diegetic sound, systematically pruning post-production background music, voiceover narration, and artificial sound effects.
Crucially, acoustic filtering evaluates sounds by their \emph{physical genesis} rather than source visibility: off-screen environmental sounds (e.g., thunder or passing vehicles) are preserved as authentic diegetic acoustics.
\end{itemize}
Because soundtrack contamination typically spans an entire video, we apply an early termination heuristic: if $10$ consecutive clips within an upload are flagged for non-diegetic audio, the remainder of that upload is terminated immediately, saving substantial inference compute.

\subsection{Camera Pose Estimation and Metric Scale Recovery (Stage 4)}
\label{app:camera}

High-fidelity interactive world modeling requires precise, metric-scaled 6-DoF camera trajectories.
For game footage, camera extrinsics and focal parameters are logged directly from engine telemetry.
For real-world and open-source footage, we develop a multi-view geometric estimation and metric scale recovery pipeline.

\begin{figure}[t]
\centering
\includegraphics[width=\linewidth]{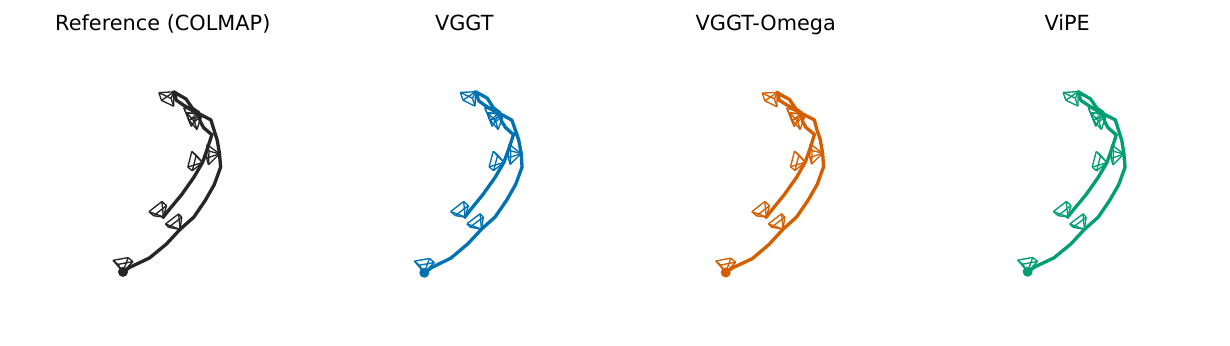}
\caption{\textbf{Single-sample camera trajectory comparison.}
From left to right: COLMAP reference, VGGT, VGGT-Omega, and ViPE.
All trajectories are aligned to the reference coordinate frame via Sim(3).
Solid lines denote camera translation paths; schematic frustums indicate camera orientations.}
\label{fig:camera-poses}
\end{figure}

\paragraph{Camera pose estimator benchmark.}
We evaluate video geometry foundation models on representative scenes from the DL3DV-10K benchmark~\citep{dl3dv}, comparing VGGT-Omega~\citep{vggtomega} against VGGT~\citep{vggt} and ViPE~\citep{vipe} using COLMAP reference poses across $32$ matched evaluation frames.
As reported in Table~\ref{tab:camera-comparison}, VGGT-Omega achieves the lowest rotation error ($0.234^\circ$) while maintaining inference efficiency comparable to VGGT ($16.09$\,s vs $187.32$\,s for ViPE).
Figure~\ref{fig:camera-poses} illustrates a representative qualitative trajectory comparison.

\begin{table}[t]
\caption{\textbf{Mean camera estimation errors and runtime across benchmark scenes.}
Translation L1 and RMSE are normalized relative to reference trajectory RMS radius after $\mathrm{Sim}(3)$ alignment; rotation error reflects global orientation alignment.
Runtime includes initialization and excludes depth recovery and window stitching.
VGGT and VGGT-Omega process $32$ evaluation frames; ViPE processes $163$--$399$ frames before subsampling.}
\label{tab:camera-comparison}
\centering
\small
\setlength{\tabcolsep}{6pt}
\begin{tabular}{@{}lcccc@{}}
\toprule
\multirow{2}{*}{Method} & \multicolumn{2}{c}{Translation (\%) $\downarrow$} & \multirow{2}{*}{Rotation ($^\circ$) $\downarrow$} & \multirow{2}{*}{Runtime (s) $\downarrow$} \\
\cmidrule(lr){2-3}
& L1 & RMSE & & \\
\midrule
VGGT~\citep{vggt} & 1.010 & 0.760 & 0.315 & \textbf{14.23} \\
\textbf{VGGT-Omega}~\citep{vggtomega} & 1.001 & 0.768 & \textbf{0.234} & 16.09 \\
ViPE~\citep{vipe} & \textbf{0.763} & \textbf{0.582} & 0.263 & 187.32 \\
\bottomrule
\end{tabular}
\end{table}

\paragraph{Metric scale recovery and temporal stitching.}
For each video clip, VGGT-Omega outputs up-to-scale relative camera extrinsics $\{\mR_t, \tilde{\vt}_t\}_{t=1}^T$ and dense relative depth maps $\tilde{D}_t \in \R^{H \times W}$.
Because interactive world models require metric physical dimensions to ground linear velocities and spatial audio inverse-square attenuation, we recover physical metric scale using monocular metric depth estimation via Depth Anything 3 (DA3)~\citep{depthanythingv3}, with Metric3D-v2~\citep{metric3dv2} serving as an architectural fallback for scenes with extreme perspective distortion.
Specifically, for keyframes sampled across the clip (at $2$\,fps), we compute the pixel-wise metric scale ratio over a confidence mask $\gM_t$ excluding sky regions and specular highlights:
\begin{equation}
r_t(u,v) = \frac{D_t^{\mathrm{metric}}(u,v)}{\tilde{D}_t(u,v)}, \quad (u,v) \in \gM_t .
\end{equation}
The global metric scale factor $s$ for the clip is computed via the robust median across all valid keyframe pixels: $s = \operatorname{median}_{(t,u,v) \in \gM} r_t(u,v)$, from which metric translations are recovered via $\vt_t = s \cdot \tilde{\vt}_t$.
When estimating extended shots across multiple consecutive 5.04-second windows, the estimator operates with a temporal sliding window overlapping by $12$ frames ($0.5$\,s).
We solve an $\mathrm{SE}(3)$ registration problem across the overlapping frames to align adjacent coordinate systems, measuring trajectory continuity by the residual camera center RMS displacement:
\begin{equation}
\mathrm{RMS}_{\mathrm{stitch}} = \sqrt{\frac{1}{K}\sum_{k=1}^K \|\hat{\vt}_{k}^{(w)} - \vt_{k}^{(w-1)}\|_2^2} ,
\end{equation}
where $\hat{\vt}_k^{(w)}$ denotes the camera position in window $w$ mapped into the reference frame of window $w-1$.

\paragraph{Kinematic plausibility gating and action derivation.}
To eliminate tracking failure artifacts and unphysical camera jumps, all estimated trajectories are validated against kinematic physical bounds:
(i)~mean velocity $\bar{v} \in [0, 40)$\,m/s,
(ii)~maximum instantaneous velocity $v_{\max} \le 50$\,m/s,
(iii)~maximum linear acceleration $a_{\max} \le 20$\,m/s$^2$, and
(iv)~cross-window stitching error $\mathrm{RMS}_{\mathrm{stitch}} \le 0.05$\,m.
No minimum-motion threshold is enforced, preserving stationary viewpoints and pure in-place rotations.
Finally, continuous relative camera motions between adjacent frames are quantized into the $81$-class discrete action vocabulary ($9\times 9$ bins of linear velocity and yaw rotation rate) detailed in Sec.~\ref{sec:action}.

\subsection{Decoupled Multimodal Captioning (Stage 4)}
\label{app:data:caption}

Conditioning an interactive audio-visual world model on a single entangled text prompt induces cross-modal hallucination during training (e.g., visual models hallucinating sound effects that are absent, or audio models generating voices when text mentions visible actors).
To prevent cross-modal contamination, we formulate a decoupled caption contract comprising three independent caption streams per clip, generated using Qwen3-Omni-30B~\citep{qwen3omni} via isolated forward passes:
\begin{itemize}[leftmargin=1.5em,itemsep=1pt,topsep=1pt]
\item \textbf{Visual caption ($V$)}: Describes setting and salient subject motion while strictly ignoring audio. Viewpoint changes and camera-induced parallax are forbidden in the text, ensuring camera dynamics remain exclusively governed by camera conditioning.
\item \textbf{Audio caption ($A$)}: Describes acoustic layers, event order, and temporal dynamics while ignoring visual content. Because captioning uses a mono downmix, directional terms (e.g., left/right) are strictly prohibited, preserving stereo panning for geometric control. Speech is treated as ambient sound without transcription.
\item \textbf{Joint audio-visual caption ($AV$)}: Explicitly grounds audible events to visible on-screen emitters, tracking cross-modal synchronization and volume changes as sound sources enter or leave the camera frustum.
\end{itemize}
This tripartite representation eliminates textual cross-talk and provides the structured textual supervision for the caption ablation in Sec.~\ref{sec:exp-main}.

\section{HelixBench and Efficiency Evaluation}
\label{app:bench}

\subsection{Case Selection}

\paragraph{Case organization.}
Each case carries a scene label and an interaction label, assigned by human annotators as content descriptions.
Per-metric applicability is decided separately from the content labels, so label counts are not metric denominators; each metric reports on its applicable subset together with the number of scored and non-scorable cases.
The benchmark contains $1{,}015$ cases; DeSync applies to the $165$ Event cases, and the spatial score to the Perspective subset.

\paragraph{Selection pipeline.}
Sources are isolated from the training corpus at the whole-video level; a source that matches training data cannot re-enter through a different time window.
Candidate windows pass technical gates: frame continuity over the full clip (scene cuts, black or frozen frames, montage effects), audio continuity and audibility (missing audio, long or edge silences, suspected fades via RMS-envelope checks), and a true-stereo gate (side/mid energy ratio $\le -45$\,dB rejected; weak-stereo cases routed to review), since duplicated mono channels carry no spatial information.

\paragraph{Spatial subset pipeline.}
The spatial subset is built by scanning full source videos: stereo energy and direction statistics are computed in 0.1-second windows, candidate anchors are proposed from audio events and visual changes under a per-interval budget, and windows of about 5 seconds are cut around anchors.
A case is admitted only if a single dominant sound source is visible and localizable; the source is bound with open-vocabulary detection using visual evidence alone, so audio direction never influences which object is selected.
Background music, rain, waves, and multi-source scenes are excluded; a reference alignment threshold selects candidates for human review, and overlapping windows are removed.

\subsection{Audio Quality}

\paragraph{KL divergence.}
Generated audio is compared with the reference audio of the corresponding case, using the first $5$ seconds of each.
PaSST~\citep{passt} produces $527$ audio-event logits per clip; applying softmax gives reference and generated distributions $p_i$ and $q_i$.
\begin{equation}
\mathrm{KL}=\frac{1}{N}\sum_{i=1}^{N}\sum_{c=1}^{527}p_{i,c}\log\frac{p_{i,c}}{q_{i,c}},\qquad N=1{,}015.
\end{equation}
The direction is reference-to-generated, $D_{\mathrm{KL}}(p_i\Vert q_i)$, with natural logarithms and equal averaging over clips.
Lower values indicate closer agreement with the reference audio-event distributions.

\paragraph{FAD.}
Fr\'echet Audio Distance~\citep{fad} compares generated and reference audio distributions in the VGGish~\citep{vggish} embedding space.
From the first $5$ seconds of each clip, five $128$-dimensional window embeddings are extracted, pooling all $5{,}075$ embeddings from the $1{,}015$ clips in each condition.
Let $(\mu_r,\Sigma_r)$ and $(\mu_g,\Sigma_g)$ denote the empirical means and covariances of the reference and generated embeddings:
\begin{equation}
\mathrm{FAD}=\|\mu_r-\mu_g\|_2^2+\operatorname{tr}\!\left(\Sigma_r+\Sigma_g-2\bigl(\Sigma_r^{1/2}\Sigma_g\Sigma_r^{1/2}\bigr)^{1/2}\right).
\end{equation}
All models use the same reference audio set; FAD is computed once per condition from the pooled window embeddings, and lower values indicate closer feature distributions.

\subsection{Cross-Modal Alignment}

\paragraph{DeSync.}
The Synchformer-based evaluator~\citep{synchformer} predicts the audio--video offset for the first and last $4.8$-second windows of each generated clip.
For each window, the offset class with the highest predicted probability is taken; the absolute values of the two offsets are averaged within a clip, then equally over clips.
The result is an estimated temporal misalignment in seconds; lower is better.

\paragraph{ImageBind.}
ImageBind~\citep{imagebind} embeds generated video and audio in a shared semantic space.
For unit-normalized video and audio embeddings $v_i$ and $a_i$ from the same generated clip, the score is the mean matched cosine similarity $\frac{1}{N}\sum_{i=1}^{N}v_i^\top a_i$ with $N=1{,}015$; higher values indicate stronger semantic agreement.

\paragraph{CLAP.}
CLAP~\citep{clap} measures cosine similarity between the generated audio embedding and the embedding of its audio-visual input caption.
The standard text encoder is used with its native $77$-token truncation, and clip scores are averaged equally over all $1{,}015$ cases.
Higher scores indicate stronger audio-caption semantic agreement.

\subsection{Spatial Audio--Visual Agreement}
\label{app:spatial}

\paragraph{Spatial three-region score.}
The score measures agreement between the horizontal position of a detected sound-source candidate in the generated video and the stereo direction of its paired generated audio.

\textit{Visual localization.} The generated video is sampled at $10$\,Hz and Grounding DINO~\citep{groundingdino} is applied with object-category queries derived from the case's video and joint audio-visual captions.
Candidate detections are associated across frames, and a target track is selected using detection confidence, temporal coverage, and visual motion; when track selection fails, the pipeline falls back to a dominant caption-matched detection in each frame.
Target selection uses visual evidence alone.
For the selected bounding box, the horizontal center $x_t$ is normalized as $d_t=2x_t/W-1$, where $W$ is the frame width.

\textit{Audio direction.} Left- and right-channel energies are computed in $100$-ms windows with a $50$-ms hop, giving the stereo energy pan $p_t=(E_{R,t}-E_{L,t})/(E_{R,t}+E_{L,t})$ with numerical stabilization; the pan sequence is interpolated to the visual sampling timestamps.
Energies are measured from the mixed audio track without source separation.
For joint audio-visual models this is the model's native generated audio; for V2A baselines it is the replacement audio synthesized for the same video, so the fixed video provides identical visual detections across audio conditions.

\textit{Three-region rule.} Visual direction is left for $d_t<-\tau_v$, right for $d_t>\tau_v$, and center otherwise, with $\tau_v=0.2$; audio direction uses the same rule on $p_t$ with $\tau_a=0.1$, equality belonging to the center region.
The visual center region therefore spans $40$--$60\%$ of image width, and the audio boundary corresponds to a channel energy ratio of $(1+\tau_a)/(1-\tau_a)\approx1.222$, or approximately $0.87$\,dB.
These tolerances stabilize left/right labels against small visual displacements and channel-energy fluctuations while retaining sensitivity to lateral deviations.

\textit{Scoring and eligibility.} For visually lateral samples, same-side audio receives $+1$, centered audio $0$, and opposite-side audio $-1$; visual-center samples are excluded.
Within each eligible clip, valid visual-side samples are averaged and multiplied by $100$; clip scores are then averaged equally over clips with at least one valid visual-side sample.
The score ranges from $-100$ to $100$, and always-centered audio scores zero.
A clip qualifies if it contains at least four consecutive valid $10$\,Hz samples; an invalid sample or timestamp gap breaks the run, and eligible clips contribute all valid samples.
The score quantifies lateral audio--visual agreement over each model's detected targets and eligible clips.

\subsection{Post-hoc Video-to-Audio Comparison}

\paragraph{Matched-video protocol.}
The dubbing comparison uses PrismAudio~\citep{prismaudio}, ThinkSound~\citep{thinksound}, and AudioX~\citep{audiox}.
All three are conditioned on the same generated videos and their original joint audio-visual captions.
Their synthesized audio replaces the original track while preserving the video stream and frame timestamps; the native audio provides the matched reference condition.
The protocol covers all $1{,}015$ test cases, with the same video and caption used across the four audio conditions of each case; evaluation follows the metric definitions and applicable subsets above.

\subsection{Spatial Threshold Sensitivity}
\label{app:spatial-sensitivity}

We evaluate sensitivity to the visual and audio thresholds defining the three spatial regions.
Tables~\ref{tab:spatial-sweep-world} and~\ref{tab:spatial-sweep-v2a} report all $20$ combinations of four visual thresholds $\tau_v\in\{0.1,0.2,1/3,0.4\}$ and five audio thresholds $\tau_a\in\{0.01,0.05,0.1,0.2,0.3\}$, reusing the same generated clips, visual detections, and audio-pan trajectories under the eligibility rule above.
The world-model sweep uses $145$ Perspective candidates, with each model contributing its eligible clips; widening the visual threshold reduces the contributing cohorts from at most $143$ clips to $113$.
The fixed-video comparison shares visual detections and the eligibility mask across all four audio conditions, with $147$ of the $150$ candidates satisfying the consecutive-sample rule and identical contributing counts for every condition.
HelixWorld ranks first among all four world models across the $20$ settings, and its native audio outperforms all three V2A replacements throughout the grid.

\begin{table}[!htbp]
\caption{\textbf{World-model sensitivity to spatial thresholds.} Results use the $145$-candidate Perspective set. Pre-trained denotes the bidirectional model after progressive condition injection. Bold scores mark the best model per setting; bold thresholds mark the main-table setting. Higher is better.}
\label{tab:spatial-sweep-world}
\centering\small
\setlength{\tabcolsep}{5pt}
\begin{tabular}{@{}ccrrrr@{}}
\toprule
$\tau_v$ & $\tau_a$ & LTX-2.3 & EchoWM & HelixWorld & Pre-trained \\
\midrule
0.10 & 0.01 & $2.16$ & $22.58$ & $\mathbf{48.69}$ & $30.14$ \\
0.10 & 0.05 & $1.96$ & $21.85$ & $\mathbf{46.52}$ & $30.34$ \\
0.10 & 0.10 & $1.55$ & $20.87$ & $\mathbf{42.15}$ & $29.30$ \\
0.10 & 0.20 & $1.16$ & $17.57$ & $\mathbf{31.65}$ & $25.91$ \\
0.10 & 0.30 & $0.90$ & $14.91$ & $\mathbf{22.94}$ & $22.80$ \\
\midrule
0.20 & 0.01 & $2.41$ & $25.92$ & $\mathbf{51.53}$ & $36.92$ \\
0.20 & 0.05 & $2.20$ & $25.30$ & $\mathbf{49.88}$ & $36.27$ \\
\textbf{0.20} & \textbf{0.10} & $1.82$ & $24.05$ & $\mathbf{46.41}$ & $33.61$ \\
0.20 & 0.20 & $1.03$ & $20.69$ & $\mathbf{35.07}$ & $28.00$ \\
0.20 & 0.30 & $0.61$ & $18.41$ & $\mathbf{25.73}$ & $24.90$ \\
\midrule
$1/3$ & 0.01 & $4.78$ & $31.07$ & $\mathbf{59.85}$ & $46.47$ \\
$1/3$ & 0.05 & $4.40$ & $31.34$ & $\mathbf{58.09}$ & $46.22$ \\
$1/3$ & 0.10 & $3.50$ & $30.38$ & $\mathbf{54.98}$ & $42.61$ \\
$1/3$ & 0.20 & $1.07$ & $24.70$ & $\mathbf{42.94}$ & $35.68$ \\
$1/3$ & 0.30 & $0.68$ & $21.75$ & $\mathbf{31.52}$ & $31.45$ \\
\midrule
0.40 & 0.01 & $5.06$ & $32.62$ & $\mathbf{62.53}$ & $46.35$ \\
0.40 & 0.05 & $4.54$ & $32.60$ & $\mathbf{61.56}$ & $46.90$ \\
0.40 & 0.10 & $4.24$ & $31.79$ & $\mathbf{58.46}$ & $43.88$ \\
0.40 & 0.20 & $1.80$ & $27.12$ & $\mathbf{46.35}$ & $37.29$ \\
0.40 & 0.30 & $1.60$ & $24.44$ & $\mathbf{34.53}$ & $32.63$ \\
\bottomrule
\end{tabular}
\end{table}

\begin{table}[!htbp]
\caption{\textbf{Native versus post-hoc audio across spatial thresholds.} All conditions use fixed HelixWorld videos from the $150$-candidate set, sharing detections and eligibility. Bold scores mark the best audio condition; bold thresholds mark the main-table setting. Higher is better.}
\label{tab:spatial-sweep-v2a}
\centering\small
\setlength{\tabcolsep}{5pt}
\begin{tabular}{@{}ccrrrr@{}}
\toprule
$\tau_v$ & $\tau_a$ & HelixWorld (native) & + AudioX & + ThinkSound & + PrismAudio \\
\midrule
0.10 & 0.01 & $\mathbf{49.32}$ & $-3.94$ & $21.72$ & $-9.94$ \\
0.10 & 0.05 & $\mathbf{47.11}$ & $-5.33$ & $21.36$ & $-8.81$ \\
0.10 & 0.10 & $\mathbf{42.79}$ & $-4.84$ & $19.04$ & $-6.69$ \\
0.10 & 0.20 & $\mathbf{32.06}$ & $-4.69$ & $14.30$ & $-4.12$ \\
0.10 & 0.30 & $\mathbf{23.07}$ & $-4.82$ & $10.47$ & $-2.23$ \\
\midrule
0.20 & 0.01 & $\mathbf{51.97}$ & $-3.73$ & $19.79$ & $-11.07$ \\
0.20 & 0.05 & $\mathbf{50.26}$ & $-4.36$ & $19.75$ & $-9.99$ \\
\textbf{0.20} & \textbf{0.10} & $\mathbf{46.88}$ & $-4.01$ & $18.05$ & $-8.04$ \\
0.20 & 0.20 & $\mathbf{35.30}$ & $-4.87$ & $13.48$ & $-4.82$ \\
0.20 & 0.30 & $\mathbf{25.70}$ & $-5.22$ & $10.07$ & $-2.44$ \\
\midrule
$1/3$ & 0.01 & $\mathbf{60.18}$ & $-3.24$ & $21.40$ & $-13.81$ \\
$1/3$ & 0.05 & $\mathbf{58.34}$ & $-1.98$ & $20.42$ & $-11.89$ \\
$1/3$ & 0.10 & $\mathbf{55.45}$ & $-2.21$ & $18.67$ & $-10.15$ \\
$1/3$ & 0.20 & $\mathbf{43.01}$ & $-4.46$ & $11.26$ & $-6.28$ \\
$1/3$ & 0.30 & $\mathbf{31.37}$ & $-5.97$ & $7.95$ & $-4.01$ \\
\midrule
0.40 & 0.01 & $\mathbf{63.28}$ & $-2.50$ & $22.17$ & $-13.41$ \\
0.40 & 0.05 & $\mathbf{62.35}$ & $-1.38$ & $21.56$ & $-11.42$ \\
0.40 & 0.10 & $\mathbf{59.32}$ & $-2.75$ & $19.37$ & $-10.39$ \\
0.40 & 0.20 & $\mathbf{46.45}$ & $-4.52$ & $12.22$ & $-7.32$ \\
0.40 & 0.30 & $\mathbf{34.46}$ & $-4.59$ & $8.55$ & $-5.09$ \\
\bottomrule
\end{tabular}
\end{table}

\subsection{Inference Efficiency}
\label{app:inference-efficiency}

\paragraph{Hardware and measurement boundaries.}
All timing runs use a single NVIDIA H800 for generation and decoding.
After two complete warm-up rollouts, we report the median of two timed runs.
Timing excludes model loading, pre-rollout text/image preparation, and MP4 encoding, but includes in-loop geometry processing, CPU offload, decoding, and frame transfer to CPU.

\paragraph{Real-time factor.}
For streaming models, steady-state RTF measures the wall-clock time after the first decoded block divided by the duration of the remaining newly generated frames.
Let $T$ be total rollout wall time, $t_1$ the first-block delivery time, $F$ the output frame count, $b$ the exclusive frame index of the first delivered block, and $f$ the output frame rate. Then
\begin{equation}
\mathrm{RTF}_{\mathrm{steady}}=\frac{T-t_1}{(F-b)/f}.
\end{equation}
The denominator counts newly generated frames after the first delivered block over the completed rollout.
Lower RTF is faster; RTF below one indicates faster-than-real-time throughput after the first block.
For the non-streaming bidirectional reference, RTF uses the complete generation-and-decoding time divided by the full generated duration.

\paragraph{Output settings.}
All models are evaluated at $768{\times}512$ resolution, using the output frame rates reported in their official repositories.

\section{Additional Trajectory Distillation Results}
\label{app:protocols}
\label{app:trajectory}
\label{app:diversity}

\subsection{Conditional Visual Diversity}

\paragraph{Inputs and features.}
Our diversity calculation follows \emph{Diversity-Preserved Distribution Matching Distillation for Fast Visual Synthesis}~\citep{dp-dmd}, adapting its per-prompt image comparison to matched video frames.
We extract DINOv3 ViT-L/16 pooled features~\citep{dinov3} and CLIP ViT-L/14 CLS features~\citep{clip} at matched times across generations of the same input, then normalize each feature vector.

\paragraph{Aggregation.}
Let $\vf_{i,t,s}$ be a unit-normalized feature for input $i$, time $t\in\mathcal T$, and generation $s\in\{1,\ldots,S\}$.
For each input we average cosine distance over generation pairs and times:
\begin{equation}
d_i=\frac{1}{|\mathcal T|\binom{S}{2}}
\sum_{t\in\mathcal T}\sum_{1\leq s<s'\leq S}
\left(1-\vf_{i,t,s}^{\top}\vf_{i,t,s'}\right).
\end{equation}
The fixed first frame is excluded.
If $\mathcal G$ is the set of sources and $\mathcal I_g$ contains the inputs from source $g$, the reported score is
\begin{equation}
D=\frac{1}{|\mathcal G|}\sum_{g\in\mathcal G}\frac{1}{|\mathcal I_g|}\sum_{i\in\mathcal I_g}d_i.
\end{equation}
Each source receives equal weight, and higher scores indicate greater conditional visual diversity.

\paragraph{Results.}
Self-forcing with online trajectory distillation loss increases DINOv3 diversity by $21.7\%$ and CLIP diversity by $13.5\%$ over DMD-only self-forcing (Table~\ref{tab:trajectory-diversity}).

\subsection{Additional Qualitative Comparisons}
\begin{figure}[!t]
\centering
\includegraphics[width=\linewidth]{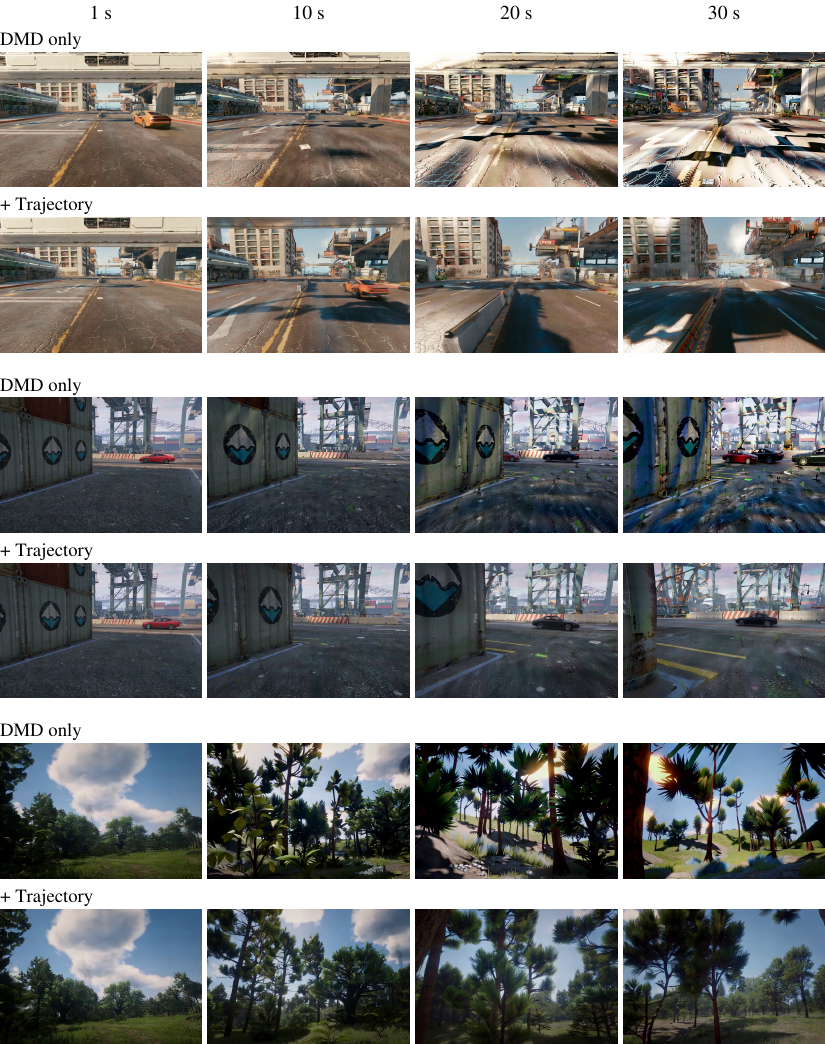}
\caption{\textbf{Additional visual comparisons of trajectory distillation.}
Each pair shows DMD only (top) and with trajectory distillation (bottom).}
\label{fig:trajectory-additional}
\end{figure}

Figure~\ref{fig:trajectory-additional} shows further examples of reduced visual drift with trajectory distillation.

\section{User Study Protocol}
\label{app:user}

\paragraph{Study design.}
Model identities are hidden, and A/B placement is randomized and balanced.
The study presents $20$ audio-enabled comparisons followed by $20$ muted-video comparisons.
With audio, we compare HelixWorld with EchoWM and three post-hoc dubbing baselines that apply AudioX, ThinkSound, or PrismAudio to the same HelixWorld video.
The muted-video comparison includes Alaya, HY-WorldPlay~1.5, LingBot~v2, SANA, and EchoWM.
We analyze $212$ audio-enabled and $190$ muted-video responses, covering $20$ scenes in each setting.

\paragraph{Evaluation criteria.}
The audio-enabled study asks about audio--visual synchrony, audio quality, spatial acoustic realism, and overall preference.
The muted-video study asks about visual quality, action following, scene following, temporal consistency, and overall preference.

\paragraph{Preference aggregation.}
Responses are summarized as a preference for HelixWorld, a tie, or a preference for the baseline.
For each model pair, a category's share is its response count divided by the total number of responses, with ties retained in the denominator.
Figure~\ref{fig:user-study} reports overall preference for each pair.
Win rates reported in the main text exclude ties.

\clearpage
\section{Additional Audio-Visual Comparisons}
\label{app:av-qualitative}

\begin{figure}[!htbp]
\centering
\includegraphics[width=\linewidth]{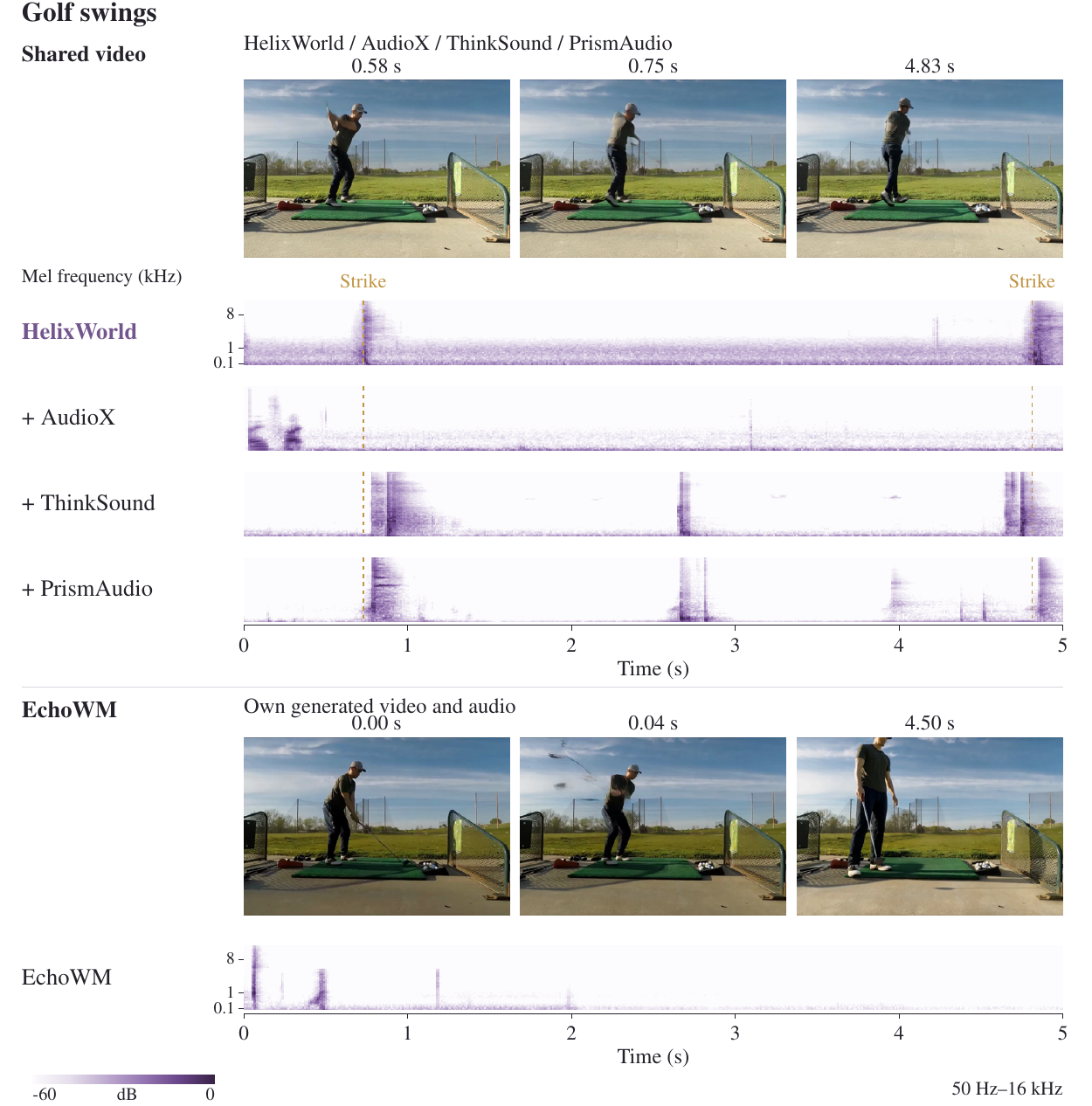}
\caption{\textbf{Audio--visual synchronization during golf swings.} Dashed lines mark the visible strikes in the shared video. HelixWorld's acoustic transients coincide with both events.}
\label{fig:av-sync-golf}
\end{figure}

\begin{figure}[!htbp]
\centering
\includegraphics[width=\linewidth]{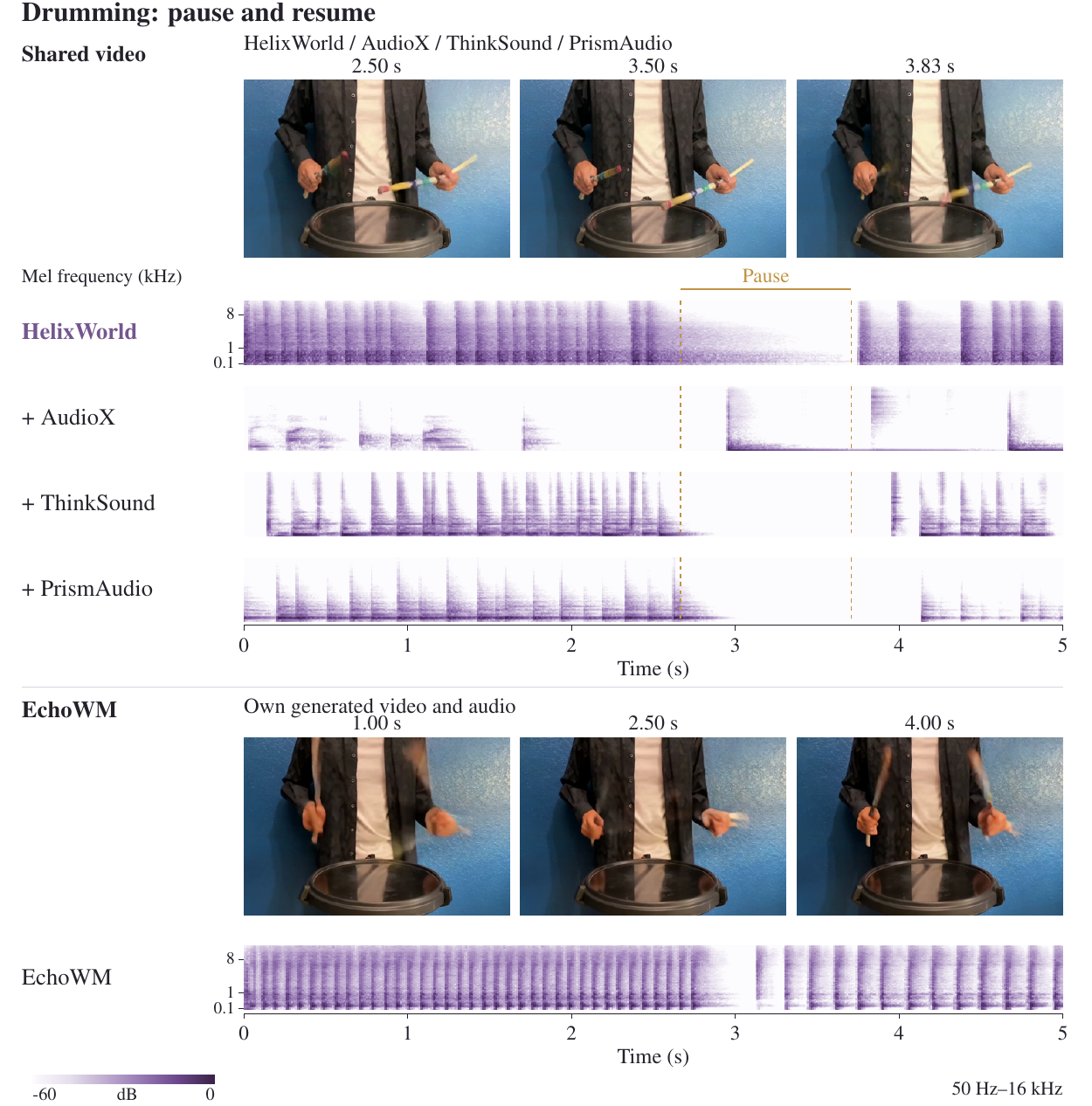}
\caption{\textbf{Audio--visual synchronization during drumming.} The marked interval is a visible pause. HelixWorld's transients subside during the pause and resume with the motion; AudioX produces a transient during the pause.}
\label{fig:av-sync-drums}
\end{figure}

\begin{figure}[!htbp]
\centering
\includegraphics[width=\linewidth]{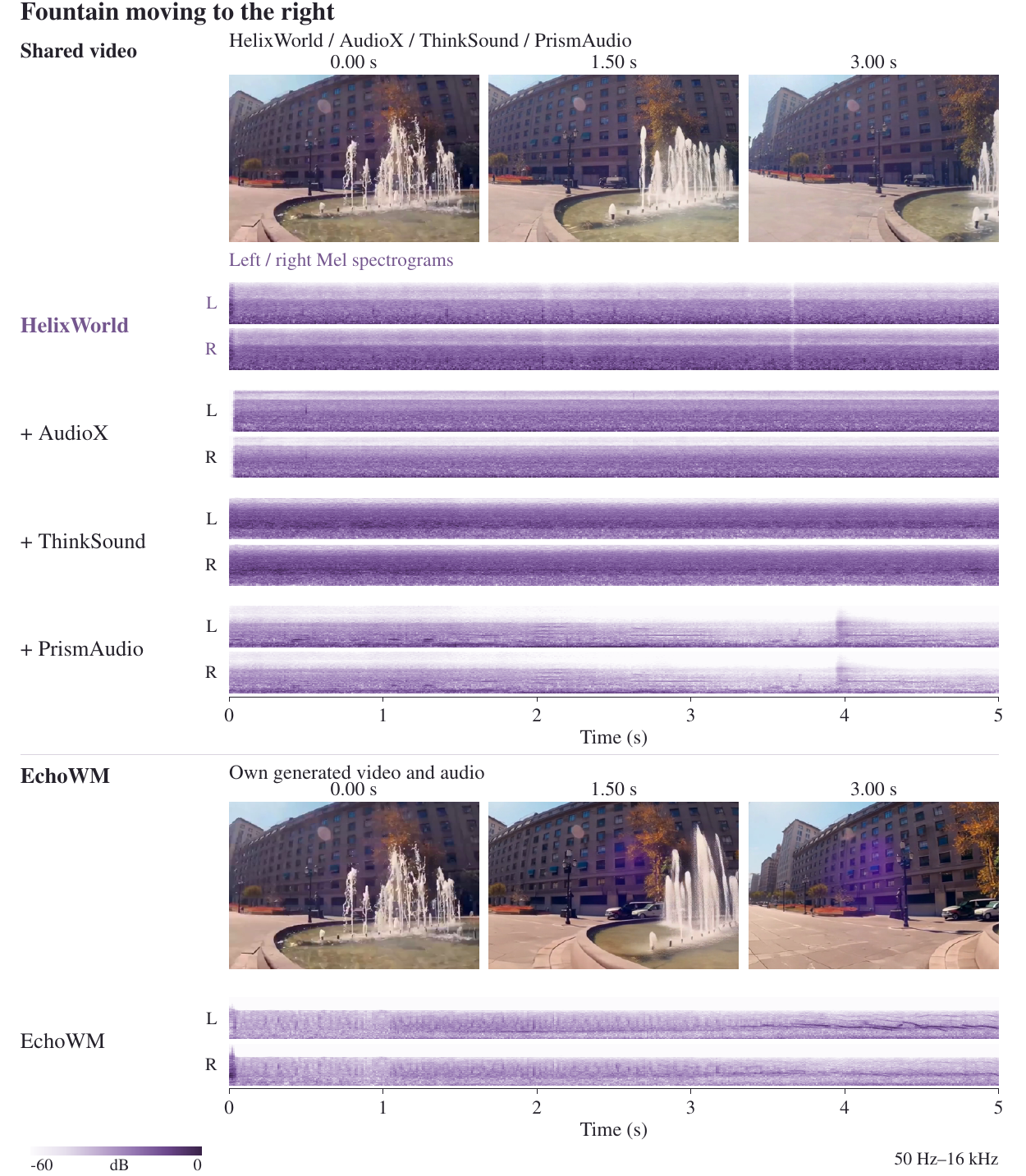}
\caption{\textbf{Stereo audio as the fountain moves rightward.} L and R denote the left and right channels. HelixWorld maintains stronger right-channel energy, consistent with the fountain's position in the video.}
\label{fig:spatial-fountain}
\end{figure}

\begin{figure}[!htbp]
\centering
\includegraphics[width=\linewidth]{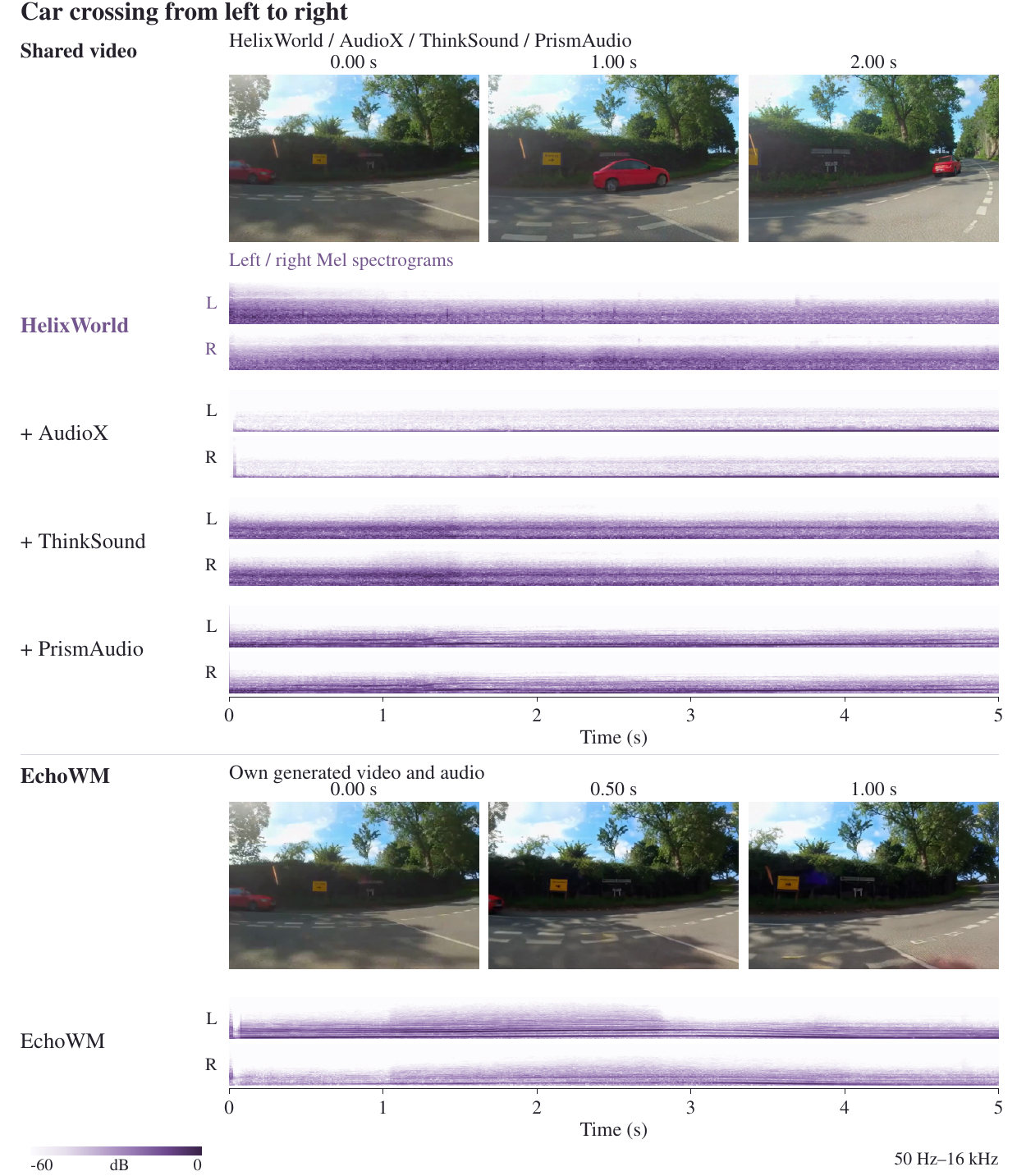}
\caption{\textbf{Stereo audio during a left-to-right vehicle pass.} HelixWorld's channel balance shifts from left to right with the car's motion. L and R denote the two audio channels.}
\label{fig:spatial-car}
\end{figure}

Figures~\ref{fig:av-sync-golf} and~\ref{fig:av-sync-drums} illustrate audio--visual synchronization; Figures~\ref{fig:spatial-fountain} and~\ref{fig:spatial-car} show stereo correspondence.
AudioX, ThinkSound, and PrismAudio use the same HelixWorld video, while EchoWM is shown with its own generated video and audio.
Mel power is normalized per recording, using a shared reference for the left and right channels.

\clearpage

\clearpage

\clearpage

\end{document}